\RequirePackage{fix-cm}
\documentclass[twocolumn]{svjour3}          
\smartqed  
\usepackage{graphicx}
\usepackage{hyperref}
\usepackage[utf8]{inputenc} 
\usepackage[T1]{fontenc}    
\usepackage{hyperref}       
\usepackage{url}            
\usepackage{booktabs}       
\usepackage{amsfonts}       
\usepackage{nicefrac}       
\usepackage{microtype}      
\usepackage{epsfig}
\usepackage{float}
\usepackage{multicol}
\usepackage{multirow}
\usepackage{amssymb}
\usepackage{amsmath}
\usepackage{wrapfig}
\usepackage[toc,page]{appendix}
\usepackage{xspace}
\usepackage{color}
\usepackage{colortbl}
\usepackage[dvipsnames, svgnames, x11names]{xcolor}
\usepackage[misc]{ifsym}
\usepackage{makecell}
\usepackage{soul}

\newcommand{\ie}{i.e}
\newcommand{\eg}{e.g}
\newcommand{\Eg}{E.g}

\def\onedot{.\xspace}
\def\eg{\emph{e.g}\onedot} 
\def\Eg{\emph{E.g}\onedot}
\def\ie{\emph{i.e}\onedot}

\def\cf{\emph{c.f}\onedot}

\def\etc{\emph{etc}\onedot}
\def\vs{\emph{vs}\onedot}

\def\etal{\emph{et al}\onedot}

\usepackage{adjustbox}
\usepackage{pifont}
\usepackage{caption}
\usepackage{array}
\usepackage{enumitem}
\newcolumntype{C}[1]{>{\centering\arraybackslash}p{#1}} 

\def \pzo {\phantom{0}} 
\newcommand{\cmark}{\ding{52}\xspace}%
\newcommand{\xmark}{\ding{56}\xspace}%
\newcommand{\xmarkg}{\textcolor{lightgray}{\ding{56}}\xspace}%
\newcommand{\pmark}{\ding{58}\xspace}%

\definecolor{lightgray}{rgb}{0.8, 0.8, 0.8}
\definecolor{lgray}{rgb}{0.66, 0.66, 0.66}

\definecolor{lgray_tab}{rgb}{0.9, 0.9, 0.9}
\definecolor{lblue_tab}{rgb}{0.717, 0.882, 0.984}

\newcommand{\blue}[1]{\textcolor[RGB]{0,0,0}{#1}}

\newcommand{\bblue}[1]{\textcolor[RGB]{0,0,0}{#1}}

\usepackage[ruled,vlined,linesnumbered]{algorithm2e}

\begin{document}
\begin{sloppypar}
\title{EATFormer: Improving Vision Transformer Inspired by Evolutionary Algorithm}

\author{Jiangning Zhang$^{1,*}$ \and
        Xiangtai Li$^{2,*}$ \and
        Yabiao Wang$^{3}$ \and
        Chengjie Wang$^{3}$ \and
        Yibo Yang$^{2}$ \and
        Yong Liu$^{1}$ \and
        Dacheng Tao$^{4}$
}

\authorrunning{Jiangning Zhang, et al.}
\institute{
  Jiangning Zhang (186368@zju.edu.cn) \\
  Xiangtai Li (lxtpku@pku.edu.cn) \\
  Yabiao Wang (caseywang@tencent.com) \\
  Chengjie Wang (jasoncjwang@tencent.com) \\
  Yibo Yang (ibo@pku.edu.cn) \\
  \Letter~Yong Liu (yongliu@iipc.zju.edu.cn) \\
  Dacheng Tao (dacheng.tao@gmail.com) \\
$^{*}$ Equally-contributed first authors. \\
$^{1}$ Institute of Cyber-Systems and Control, Advanced Perception on Robotics and Intelligent Learning Lab (APRIL), Zhejiang University, China. \\
$^{2}$ School of Artificial Intelligence, Key Laboratory of Machine Perception (MOE), Peking University, China. \\
$^{3}$ Youtu Lab, Tencent, China. \\
$^{4}$ School of Computer Science, Faculty of Engineering, The University of Sydney, Darlington, NSW 2008, Australia.
}

\date{Received: date / Accepted: date}
\maketitle

\begin{abstract} \label{section:abs}
  Motivated by biological evolution, this paper explains the rationality of Vision Transformer by analogy with the proven practical Evolutionary Algorithm (EA) and derives that both have consistent mathematical formulation. Then inspired by effective EA variants, we propose a novel pyramid EATFormer backbone that only contains the proposed \emph{EA-based Transformer} (EAT) block, which consists of three residual parts, \ie, \emph{Multi-Scale Region Aggregation} (MSRA), \emph{Global and Local Interaction} (GLI), and \emph{Feed-Forward Network} (FFN) modules, to model multi-scale, interactive, and individual information separately. Moreover, we design a \emph{Task-Related Head} (TRH) docked with transformer backbone to complete final information fusion more flexibly and \emph{improve} a \emph{Modulated Deformable MSA} (MD-MSA) to dynamically model irregular locations. Massive quantitative and quantitative experiments on image classification, downstream tasks, and explanatory experiments demonstrate the effectiveness and superiority of our approach over State-Of-The-Art (SOTA) methods. \Eg, our Mobile (1.8M), Tiny (6.1M), Small (24.3M), and Base (49.0M) models achieve 69.4, 78.4, 83.1, and 83.9 Top-1 only trained on ImageNet-1K with naive training recipe; EATFormer-Tiny/Small/Base armed Mask-R-CNN obtain 45.4/47.4/49.0 box AP and 41.4/42.9/44.2 mask AP on COCO detection, surpassing contemporary MPViT-T, Swin-T, and Swin-S by 0.6/1.4/0.5 box AP and 0.4/1.3/0.9 mask AP separately with less FLOPs; Our EATFormer-Small/Base achieve 47.3/49.3 mIoU on ADE20K by Upernet that exceeds Swin-T/S by 2.8/1.7. Code is available at \url{https://github.com/zhangzjn/EATFormer}.
  
\end{abstract}

\section{Introduction} \label{section:intro}
Since Vaswani~\etal~\cite{attn} introduce the Transformer that achieves outstanding success in the machine translation task, many improvements have been made over this structure~\cite{attn_improve3,attn_improve4,attn_bert}. Subsequently, Alexey~\etal~\cite{attn_vit} firstly introduce Transformer to the computer vision field and propose a novel ViT model that successfully sparks a new wave of research besides conventional CNN-based vision models. Recently, many excellent vision transformer models~\cite{attn_pvt,attn_swin,attn_xcit,attn_metaformer,attn_nat,attn_mpvit,attn_uniformer} have been proposed and have achieved great success in the field of many vision tasks. Currently, many attempts have been made to explain and improve the Transformer structure from different perspectives~\cite{theory_more_inductive_bias,theory_more1,theory1,theory2,theory3,attn_convit,theory_more2,theory_more3,attattr}, while continuing research is still needed. Most current models generally migrate the structural design of CNN, and they are experimentally conducted to verify the effectiveness of modules or improvements, which lacks explanations about \emph{why improved Transformer approaches work}~\cite{attattr,attn_metaformer,attn_notattention}.

\begin{figure*}[htp]
  \centering
  \includegraphics[width=1.0\linewidth]{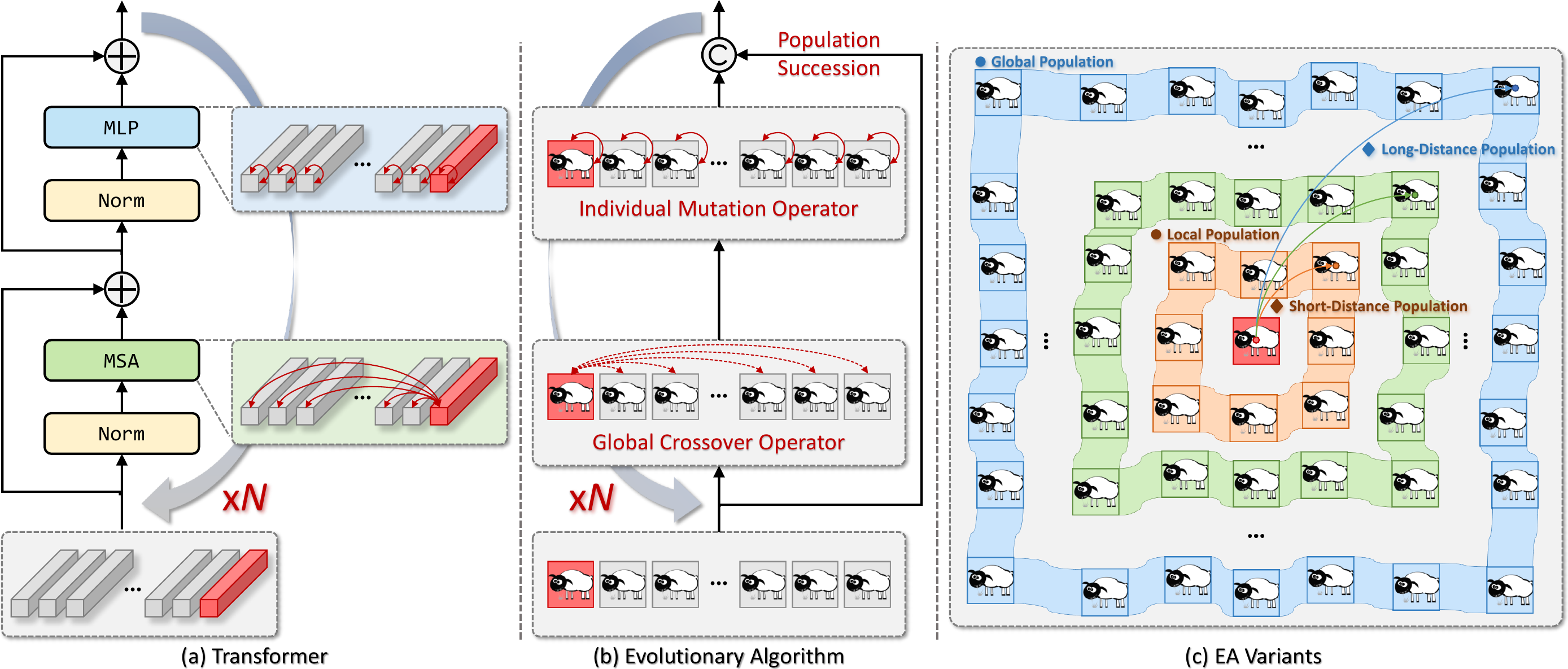}
  \caption{Comparisons of structural analogousness between (a)-Transformer module and (b)-evolutionary algorithm, where they have analogous concepts of 1) individual definition (Token Embedding \vs Individual), 2) global information interaction (MSA \vs Crossover), 3) individual feature enhancement (FFN \vs Mutation), 4) feature inheritance (Skip Connection \vs Population Succession), \etc (c)-Intuitive illustration of some EA variants.}
  \label{fig:motivation}
\end{figure*}

Inspired by biological population evolution, we explain the rationality of Transformer by analogy with the proven effective, stable, and robust Evolutionary Algorithm (EA) in this article, which has been widely used in many practical applications. 
We observe that the procedure of the Transformer (\textit{abbr.}, TR) has similar attributes to the naive EA through analogical analysis in Figure~\ref{fig:motivation} (a)-TR and (b)-EA:
\begin{description}[labelindent=0pt, leftmargin=0pt]
  \item[\textbf{1)}] In terms of data format, TR processes patch embeddings while EA evolutes individuals, both of them have the same data formats and necessary initialization. 
  \item[\textbf{2)}] In terms of optimization objective, TR aims to obtain an optimal vector representation that fuses global information through multiple layers (denoted as \textcolor[RGB]{192,0,0}{\textbf{x\emph{N}}} in Figure~\ref{fig:motivation}), while EA focuses on getting the best individual globally through multiple iterations.
  \item[\textbf{3)}] In terms of component, Multi-head Self-Attention (MSA) in TR aims to enrich patch embeddings through global information communication densely, while crossover operation in EA plays the role of interacting global individuals sparsely. Also, Feed-Forward Network (FFN) in TR enhances every single embedding for all spatial positions, which is similar to Mutation in EA that evolves each individual of the whole population. 
  \item[\textbf{4)}] Furthermore, we deduce the mathematical characterization of crossover and mutation operators in EA (\cf, Equations~\ref{eq:crossover},\ref{eq:mutation}) and find that they \emph{have the same mathematical formulations} as MSA and FFN in TR (\cf, Equations~\ref{eq:msa},\ref{eq:ffn}), respectively. 
\end{description}

In addition to the above basic analogies between naive Transformer and EA, we explore further to improve the current vision transformer by leveraging other domain knowledge of EA variants. Without losing generality, we only study the widely used and effective EA methods that could inspire us to improve Transformer. They can be mainly divided into the following categories:
\begin{description}[labelindent=0pt, leftmargin=0pt]
  \item[\textbf{1)}] \emph{Global and Local populations} inspired \emph{simultaneously global and local modeling}. In contrast to naive EA that only models global interaction, local search-based EA variants focus on finding a better individual in its neighborhood~\cite{lea1,lea2,lea4} that is more efficient without associating the global search space. Furthermore, Moscato~\etal~\cite{mea1} firstly propose the Memetic Evolutionary Algorithm (MEA) that introduces a local search process for converging high-quality solutions faster than conventional evolutionary counterparts, and intuitive illustration can be viewed in Figure~\ref{fig:motivation}-(c). For a particular individual (\ie, center sheep with red background), naive EA only contains \textcolor[RGB]{45,117,182}{\ding{108} \textbf{Global Population}} concept, while \textcolor[RGB]{132,59,12}{\ding{108} \textbf{Local Population}} idea enables the model to focus on more relevant individuals. Inspired by those EA variants, we revisit the global MSA part and \emph{improve} a novel \emph{Global and Local Interaction} (GLI) module, which is designed as a parallel structure that employs an extra local operation beside the global operation, \ie, introducing inductive bias and locality in MSA. The former is used to mine more relevant local information, while the latter aims to model global cue interactions. Considering that the spatial relationship among real individuals will not be as horizontal and vertical as the image, we further propose a Modulated Deformable MSA (MD-MSA) to dynamically model irregular locations, which could focus on more informative reorganizational regions.
  \item[\textbf{2)}] \emph{Multi-population} inspired \emph{multi-scale information aggregation}. Some works~\cite{lpea2,lpea4} introduce multi-population evolutionary algorithm to solve the optimization problems, which adopts different searching regions to more efficiently enhance the diversity of individuals and can obtain a better model performance significantly. As shown in Figure~\ref{fig:motivation}-(c), \textcolor[RGB]{45,117,182}{\ding{117} \textbf{Long-Distance Population}} could supplement more diverse and richer cues, while \textcolor[RGB]{132,59,12}{\ding{117} \textbf{Short-Distance Population}} focuses on providing general evolutionary features. Analogously, this idea inspires us to design a \emph{Multi-Scale Region Aggregation} (MSRA) module that aggregates information from different receptive fields for vision transformer, which could integrate more expressive features from different resolutions before feeding them into the next module. 
  \item[\textbf{3)}] \emph{Dynamic population} inspired \emph{pyramid architecture design}. The works~\cite{alpha2,alpha3,alpha4} investigate jDEdynNP-F algorithm with a dynamic population reduction scheme that significantly improves the effectiveness and accelerates the convergence of the model, which is similar to pyramid-alike improvements of some current vision transformers~\cite{attn_pvt2,attn_twins,attn_swin2,attn_nat,attn_metaformer}. Analogously, we extend our previous columnar-alike work~\cite{eat} to a pyramid structure like PVT~\cite{attn_pvt}, which significantly boosts the performance for many vision tasks. 
  \item[\textbf{4)}] \emph{Self-adapted parameters} inspired \emph{weighted operation mixing}. Brest~\etal~\cite{alpha1} propose an adaptation mechanism to control different optimization processes for better results, and some memetic EAs~\cite{mea1,mea2,mea3} own a similar concept of search intensity to balance the global and local calculation. This encourages us to learn appropriate weights for different operations, which can increase the performance and be more interpretable.
  \item[\textbf{5)}] \emph{Multi-objective EA} inspired \emph{task-related feature merging}. Current TR-based vision models would initialize different tokens for different tasks~\cite{attn_deit} (\eg, classification and distillation) or use the pooling operation to obtain global representation~\cite{attn_swin}. However, both manners suffer from potentially incompatibility: the former treats the task token and image patches coequally that is unreasonable, and the calculation of each layer will slightly increase the amount of calculation ($O(n^2)$ to $O((n+1)^2)$), while the latter uses only one pooling result for multiple tasks that could potentially damage the model accuracy. Inspired by multi-objective EAs~\cite{moea1,moea2,moea3} that find a set of solutions for different targets, we design a \emph{Task-Related Head} (TRH) docked with transformer backbone to complete final information fusion, which is elegant and flexible for different tasks learning.
\end{description}

\begin{figure}[tp]
  \centering
  \includegraphics[width=1.0\linewidth]{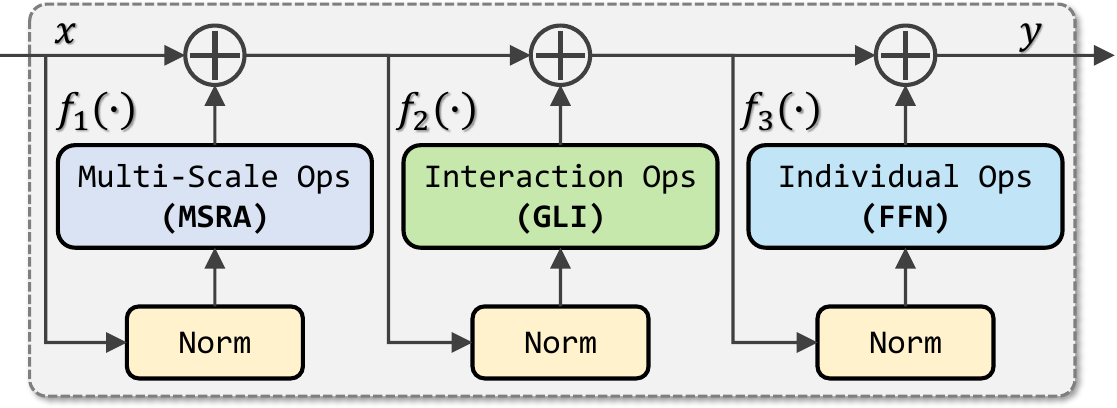}
  \caption{Paradigm of the proposed basic EAT Block, which contains three $y = f(x) + x$ residuals to model: (a) multi-scale information aggregation, (b) feature interactions among tokens, and (c) individual enhancement.}
  \label{fig:paradigm}
\end{figure}

\begin{figure}[tp]
  \centering
  \includegraphics[width=1.0\linewidth]{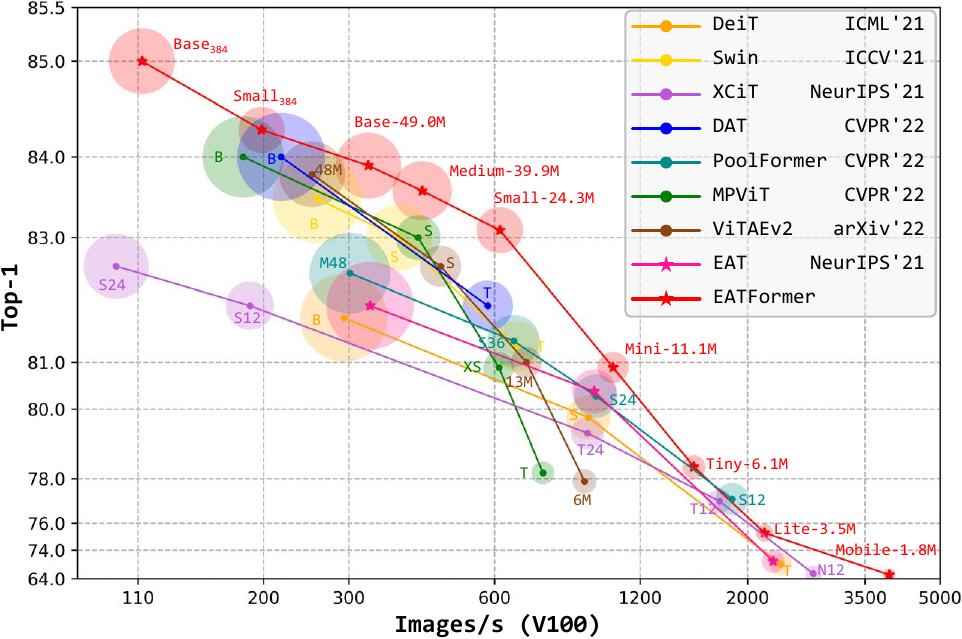}
  \caption{Comparison with SOTAs in terms of Top-1 \vs GPU throughput. All models are trained only with ImageNet-1K~\cite{imagenet} dataset in 224$\times$224, and the radius represents the relative number of parameters.}
  \label{fig:sotas}
\end{figure}

Based on the above analyses, we improve our columnar EAT model~\cite{eat} to a pyramid EA-inspired Transformer (EATFormer) that achieves a new SOTA result. Figure~\ref{fig:sotas} illustrates intuitive comparisons with SOTAs under GPU throughput, Top-1, and the number of parameters evaluation indexes, where our smallest EATFormer-Mobile obtains 69.4 Top-1 with 3,926 throughput under one V100 GPU, and the EATFormer-Base achieves 83.9 Top-1 with only 49.0M parameters. Specifically, we make the following four contributions compared with the previous conference work:
\begin{description}[labelindent=0pt, leftmargin=0pt]
  \item[\ding{228}] In theory, we enrich evolutionary explanations for \emph{the rationality of Vision Transformer} and derive a consistent mathematical formulation with evolutionary algorithm.
  \item[\ding{228}] On framework, we propose a novel basic \emph{EA-based Transformer} (EAT) block (shown in Figure~\ref{fig:paradigm}) that consists of three residual parts to model multi-scale, interactive, and individual information, respectively, which is stacked to form our proposed pyramid EATFormer.
  \item[\ding{228}] For method, inspired by effective EA variants, we analogously design: \emph{1)} Global and Local Interaction module, \emph{2)} Multi-Scale Region Aggregation module, \emph{3)} Task-Related Head module, and \emph{4)} Modulated Deformable MSA module to improve effectiveness and usability of our EATFormer.
  \item[\ding{228}] Massive experiments on classification, object detection, and semantic segmentation tasks demonstrate the superiority and efficiency of our approach, while ablation and explanatory experiments further prove the efficacy of EATFormer and its components.
\end{description}

\section{Related Work} \label{section:related}
\subsection{Evolution Algorithms}
Evolution algorithm (EA) is a subset of evolutionary computation in computational intelligence that belongs to modern heuristics, and it serves as an essential umbrella term to describe population-based optimization and search techniques in the last 50 years~\cite{ea1,ea2,bartz2014evolutionary}. Inspired by biological evolution, general EAs mainly contain reproduction, crossover, mutation, and selection steps, which have been proven effective and stable in many application scenarios~\cite{mea1,mea2}, and a series of improved EA approaches have been advanced in succession. 
Differential Evolution (DE) developed in 1995~\cite{dea1} is arguably one of the most competitive improved EA that significantly advances the global modeling capability~\cite{dea2,dea3}. The core idea of DE is introducing a complete differential concept to the conventional EA, which differentiates and scales two individuals in the same population and interacts with the third individual to generate a new individual. 
In contrast to the category mentioned above, local search-based EAs aim to find a solution that is as good as or better than all other solutions in its neighborhood~\cite{lea1,lea2,lea4}. This thought is more efficient than global search in that a solution can quickly be verified as a local optimum without associating the global search space. However, the locality-aware operation will restrict the ability of global modeling that could lead to suboptimal results in some scenarios, so some researchers attempt to fuse both above modeling manners. 
Moscato~\etal~\cite{mea1} firstly propose the Memetic Evolutionary Algorithm (MEA) in 1989 that applies a local search process to refine solutions for hard problems, which could converge to high-quality solutions more efficiently than conventional evolutionary counterparts. In detail, this variant is a particular global-local search hybrid: the global character is given by the traditional EA, while the local aspect is mainly performed through constructive methods and intelligent local search heuristics~\cite{mea2}. 
Analogously, some later works~\cite{lpea2,lpea4} introduce a multi-population evolutionary algorithm to solve the constrained function optimization problems relatively efficiently, which adopts different searching regions to enhance the diversity of individuals that improves the model ability dramatically. This strategy inspires us to design a basic feature extraction module for vision transformer: whether a similar multi-scale manner can be adopted to enhance model expressiveness. 
Furthermore, Brest~\etal~\cite{alpha1} propose an adaptation mechanism on the control parameters $CR$ and $F$ for crossover and mutation operations associated with DE, where adapted parameters are applied to different optimization processes for obtaining better results. Remarkably, those MEAs~\cite{mea1,mea2,mea3} mentioned above own a similar concept of search intensity to balance the global and local calculation. 
Subsequent work~\cite{alpha2} investigates jDEdynNP-F algorithm with a dynamic population reduction scheme, where the population size of the next generation is equal to half the previous population size. This strategy significantly improves the effectiveness and accelerates the convergence of the model that is consistently illustrated by works~\cite{alpha3,alpha4}. Furthermore, some literatures~\cite{bio1,bio2,bio3} suggest that there are hierarchical structures of V1, V2, V4, and inferotemporal cortex in the evolutionary brain, which have ordered interconnection among them in the form of both feed-forward and feedback connections. 
Moreover, researchers~\cite{moea1,moea2,moea3} study multi-objective EAs to find optimal trade-offs to get a set of solutions for different targets.

Inspired by the aforementioned EA variants that introduce various and valid concepts for optimization, we explain and improve the naive transformer structure by conceptual analogy in the paper, where a novel and potent EATFormer with pyramid architecture, multi-scale region aggregation, and global-local modeling is hand-designed. Furthermore, a plug-and-play task-related head module is developed to solve different targets separately and improve model performance.

\subsection{Vision Transformers}
Since Transformer structure achieves significant progress for machine translation task~\cite{attn}, many improved language models~\cite{attn_elmo,attn_bert,attn_gpt1,attn_gpt2,attn_gpt3} are proposed and obtain great achievements, and some later works~\cite{attn_improve1,attn_improve2,attn_improve3,attn_improve4,attn_improve5,wang2021evolving} advance the basic transformer module for better efficiency. Inspired by the success of Transformer in NLP and the rapid improvement of computing power, Alexey~\etal~\cite{attn_vit} propose a novel ViT that firstly introduces the transformer to vision classification and sparks a new wave of research besides conventional CNN-based vision models. Subsequently, many excellent vision transformer models are proposed, and they can mainly be divided into two categories, \ie, pure and hybrid vision transformers. 
\emph{The former} only contains transformer module without CNN-based layers, and early works\cite{attn_deit,attn_deepvit,attn_tnt,attn_cait,attn_t2tvit,attn_cpvt,attn_xcit} follow columnar structure of original ViT. Typically, DeiT~\cite{attn_deit} propose an efficient training recipe to moderate the dependence on large datasets, DeepViT~\cite{attn_deepvit} and CaiT~\cite{attn_cait} focus on fast performance saturation when scaling ViT to be deeper, and TNT~\cite{attn_tnt} divide local patches into smaller patches for fine-grained modeling. Furthermore, researchers~\cite{attn_pvt,attn_swin,attn_shuffle,attn_halonets,attn_twins,attn_swin2,attn_nat,attn_dpt,attn_metaformer,attn_SSA} advance ViT to pyramid structure that is more powerful and suitable for dense prediction. PVT~\cite{attn_pvt} leverages a non-overlapping patch partition to reduce feature size, while Swin~\cite{attn_swin} utilizes a shifted window scheme to alternately model in-window and cross-window connection. 
\emph{The latter} incorporates the idea of convolution that owns natural inductive bias of locality and translation invariance, and this kind of combination dramatically improves the model effect. 
Specifically, Srinivas~\etal~\cite{attn_botnet} advance CNN-based models by replacing the convolution of the bottleneck block with the MSA structure. Later researches~\cite{attn_localvit,attn_ceit,attn_convit,attn_vitae} introduce convolution designs into columnar visual transformers, while works~\cite{attn_lvt,attn_volo,attn_crossformer,attn_container,attn_uniformer,attn_coat,attn_cvt,attn_pvt2,attn_vitae2} fuse convolution structures into pyramid structure or use CNN-based backbone on early stages, which has obvious advantages over pure transformer models. Moreover, some researchers design hybrid models from the parallel perspective of convolution and transformer~\cite{attn_coat,attn_mobileformer,attn_mixformer}, while Xia~\etal~\cite{attn_vit} introduce a deformable idea~\cite{dcn1} to MSA module and obtain a boost on Swin~\cite{attn_swin}. Recently, MPViT~\cite{attn_mpvit} explores multi-scale patch embedding and multi-path structure that enable both fine and coarse feature representations simultaneously. 
Benefiting from advances in basic vision transformer models, many task-specific models are proposed and achieve significant progress in down-stream vision tasks, \eg, 
object detection~\cite{down_detr,down_deformabledetr,down_YOLOS,down_pix2seq}, 
semantic segmentation~\cite{down_setr,down_transunet,down_medt,down_segformer,down_hrformer,down_maskformer,down_mask2former}, 
generative adversarial network~\cite{down_transgan,down_fat,down_gat}, 
low-level vision~\cite{down_ipt,down_swinir,down_restormer}, 
video understanding~\cite{down_vtn,down_timesformer,down_STRM,down_bevt,down_LSTR}, 
self-supervised learning~\cite{down_sit,down_mocov3,down_fdsl,down_mae,down_beit,down_dino,down_peco,down_cae,down_simmim,down_maskfeat,down_data2vec,down_convmae}, 
neural architecture search~\cite{down_hat,down_bossnas,down_glit,down_autoformer,down_s3}, \etc
Inspired by practical improvements in EA variants, this work migrates them to Transformer improvement and designs a powerful visual model with higher precision and efficiency than contemporary works. Also, thanks to the elaborate analogical design, the proposed EATFormer in this paper is highly explanatory.

\vspace{-1.0em}
\subsection{Explanatory of Transformers}
\vspace{-0.5em}
Transformer-based models have achieved remarkable results on CV tasks, leading us to question why Transformer works so well, even better than Convolutional Neural Networks (CNN). Many efforts have done to answer this question. 
Goyal~\etal~\cite{theory_more_inductive_bias} point out that studying the kind of inductive biases that humans and animals exploit could help inspire AI research and neuroscience theories. 
Pan~\etal~\cite{theory_more1} show a strong underlying relation between convolution and self-attention operations. Jean-Baptiste~\etal~\cite{theory1} prove that a multi-head self-attention layer with a sufficient number of heads is at least as expressive as any convolutional layer, while Raghu~\etal~\cite{theory2} find striking differences between ViT and CNN on image classification. 
Kunchang~\etal~\cite{attn_uniformer} seamlessly integrate the merits of convolution and self-attention in a concise transformer format, while ConVit~\cite{attn_convit} combines the strengths of both CNN/Transformer architectures to introduce gated positional self-attention. Introducing local CNN into Transformer is followed by many subsequent works~\cite{attn_cpvt,attn_uniformer,cmt,attn_vitae2,mobilevit,edgenext}. Furthermore, Juhong~\etal~\cite{theory3} take a biologically inspired approach and explore modeling peripheral vision by incorporating peripheral position encoding to the multi-head self-attention layers in Transformer. 
Besides, some works explore the relation between Transformer and other models, \eg, Katharopoulos~\etal~\cite{theory_more_rnn} reveals their relationship of Transformer to recurrent neural networks, and Kim~\etal~\cite{theory_more_gnn} prove that Transformer is theoretically at least as expressive as an invariant graph network composed of equivariant linear layers. 
Moreover, Bhojanapalli~\etal~\cite{theory_more2} find that ViT models are at least as robust as the ResNet counterparts on a broad range of perturbations when pre-trained with a sufficient amount of data. 
Hao~\etal~\cite{attattr} propose a self-attention attribution method to interpret the information interactions inside Transformer, while Liu~\etal~\cite{theory_more3} propose an actionable diagnostic methodology to measure the consistency between explanation weights and the impact polarity for attention-based models. Dong~\etal~\cite{attn_notattention} find that MLP stops the output from degeneration, and removing MSA in Transformer would also significantly damage the performance. 
Recently, Qiang~\etal~\cite{theory_more4} propose a novel Transformer explanation technique via attentive class activation tokens by leveraging encoded features, gradients, and attention weights to generate a faithful and confident explanation. Xu~\etal~\cite{theory_more5} propose a new way to visualize the model by firstly computing attention scores based on attribution and then propagating these attention scores through the layers. 
Works~\cite{attn_metaformer,zhang2023rethinking} demonstrate that the general architecture of the Transformers is more essential to performance rather than the specific token mixer module. 
The above work explores the interpretation of Transformer from a variety of perspectives. At the same time, we will provide another explanation from the perspective of evolutionary algorithms and design a robust model to perform multiple CV tasks.

\section{Preliminary Transformer} \label{section:pre}
The vision transformer generally refers to the encoder part of the original transformer structure, which consists of Multi-head Self-Attention layer (MSA), Feed-Forward Network (FFN), Layer Normalization (LN), and Residual Connection (RC). Given the input feature maps $\boldsymbol{X}_{img} \in \mathbb{R}^{C \times H \times W}$, $\operatorname{Img2Seq}$ operation firstly flattens it to a 1D sequence $\boldsymbol{X}_{seq} \in \mathbb{R}^{C \times N}$ that complies with standard NLP format, denoted as: $\boldsymbol{X}_{seq} = \operatorname{Img2Seq}(\boldsymbol{X}_{img})$. 

\begin{description}
  \item[\ding{112}] \textbf{MSA} fuses several SA operations to process $\boldsymbol{QKV}$ that jointly attend to information in different representation subspaces. Specifically, LN solved $\boldsymbol{X}_{seq}$ goes through linear layers to obtain projected queries ($\boldsymbol{Q}$), keys ($\boldsymbol{K}$) and values ($\boldsymbol{V}$) presentations, formulated as:
  \begin{equation}
    \resizebox{.9\hsize}{!}{$
    \begin{aligned}
      \operatorname{MSA}(\boldsymbol{X}_{seq})  &=  \left(\bigoplus_{h=1}^{H} \boldsymbol{X}_{h}\right) \boldsymbol{W}^{O}, \\
      \text {where}~ \boldsymbol{X}_{h}  &=  \operatorname{Attention} \left( \boldsymbol{X}_{seq} \boldsymbol{W}_{h}^{Q}, \boldsymbol{X}_{seq} \boldsymbol{W}_{h}^{K}, \boldsymbol{X}_{seq} \boldsymbol{W}_{h}^{V} \right) \\
                                                &=  \operatorname{Softmax} \left( \left[ \frac{(\boldsymbol{X}_{seq} \boldsymbol{W}_{h}^{Q}) (\boldsymbol{X}_{seq} \boldsymbol{W}_{h}^{K})^{T}}{\sqrt{d_{k}}} \right] \right) \boldsymbol{X}_{seq} \boldsymbol{W}_{h}^{V} \\ 
                                                &=  \operatorname{Softmax} \left( \left[ \frac{\boldsymbol{Q}_{h} \boldsymbol{K}_{h}^{T}}{\sqrt{d_{k}}}\right] \right) \boldsymbol{V}_{h} \\ 
                                                &=  \boldsymbol{A}_{h} \boldsymbol{V}_{h} ,
    \end{aligned}
    $}
    \label{eq:tr_msa}
  \end{equation}
  where $d_m$ is the input dimension, while $d_q$, $d_k$, and $d_v$ are hidden dimensions of the corresponding projection subspace, and generally $d_q$ equals $d_k$; $\mathrm{h}$ is the head number; $\boldsymbol{W}_{h}^{Q} \in \mathbb{R}^{d_{m} \times d_{q}}$, $\boldsymbol{W}_{h}^{K} \in \mathbb{R}^{d_{m} \times d_{k}}$, and $\boldsymbol{W}_{h}^{V} \in \mathbb{R}^{d_{m} \times d_{v}}$ are parameter matrices for $\boldsymbol{QKV}$, respectively; $\boldsymbol{W}^{O} \in \mathbb{R}^{h d_{v} \times d_{m}}$ maps each head feature $\boldsymbol{X}_{h}$ to the output; $\oplus$ means concatenation operation; $\boldsymbol{A}_{h} \in \mathbb{R}^{l \times l}$ is the attention matrix of $h$-th head.
  \item[\ding{112}] \textbf{FFN} consists of two cascaded linear transformations with a ReLU activation in between:
  \begin{equation}
    \begin{aligned}
      \mathrm{FFN}(\boldsymbol{X}_{seq})=\max \left(0, \boldsymbol{X}_{seq} \boldsymbol{W}_{1} + \boldsymbol{b}_{1}\right) \boldsymbol{W}_{2} + \boldsymbol{b}_{2},
    \end{aligned}
    \label{eq:tr_ffn}
  \end{equation}
  where $\boldsymbol{W}_1$ and $\boldsymbol{W}_2$ are weights of two linear layers, while $\boldsymbol{b}_1$ and $\boldsymbol{b}_2$ are corresponding biases. 
  
  \item[\ding{112}] \textbf{LN} is applied before each layer of MSA and FFN, and the transformed $\hat{\boldsymbol{X}}_{seq}$ is calculated by:
  \begin{equation}
    \begin{aligned}
      \hat{\boldsymbol{X}}_{seq} = \boldsymbol{X}_{seq} + [\text{MSA}~|~\text{FFN}](\text{LN}(\boldsymbol{X}_{seq})).
    \end{aligned}
  \end{equation}
  \label{eq:tr_ln}
\end{description}
Finally, reversed $\operatorname{Seq2Img}$ operation \emph{reshapes} the enhanced $\hat{\boldsymbol{X}}_{seq}$ back to 2D feature maps, denoted as: $\hat{\boldsymbol{X}}_{img} = \operatorname{Seq2Img}(\hat{\boldsymbol{X}}_{seq})$. 

\section{EA-Inspired Vision Transformer} \label{section:method}
In this section, we expand the relationship among operators in naive EA and modules in naive Transformer, and consistent mathematical formulations for each conceptual pair can be derived, revealing evolutionary explanations for \emph{the rationality of Vision Transformer structure}. Inspired by the core ideas of some effective EA variants, we deduce them into transformer architecture design and improve a \emph{mighty} pyramid EATFormer over the previous columnar model.

\subsection{Evolutionary Explanation of Transformer}
As aforementioned in Figure~\ref{fig:motivation}, the Transformer block has conceptually similar sub-modules analogously to evolutionary algorithm. Basically, Transformer inputs a sequence of patch tokens while EA evolutes a population that consists of many individuals. Both of which have the consistent vector format and necessary initialization. In order to facilitate the subsequent analogy and formula derivation, we symbolize the patch token (individual) as $\boldsymbol{x}_{i}=[ x_{i,1}, x_{i,2}, \dots, x_{i,D} ]$, where $i$ and $D$ indicate data order and dimension, respectively. Define $L$ as the sequence length, the sequence (population) can be denoted as $\boldsymbol{X} = [\boldsymbol{x}_{1}, \boldsymbol{x}_{2}, \dots, \boldsymbol{x}_{L}]^{\mathrm{T}}$. The specific relationship analyses of different components are as follows:

\begin{description}[labelindent=0pt, leftmargin=0pt]
\item[\ding{80}] \textbf{Crossover Operator} \vs \textbf{MSA Module}. \\
\emph{For the crossover operator of EA}, it aims at creating new individuals by combining parts of other individuals. For an individual $\boldsymbol{x}_{i}$ specifically, the operator will randomly pick another individual $\boldsymbol{x}_{j}=[ x_{j,1}, x_{j,2}, \dots, x_{j,D} ] (1 \leq j \leq L)$ in the global population and randomly replaces features of $\boldsymbol{x}_{i}$ with $\boldsymbol{x}_{j}$ to form the new individual $\hat{\boldsymbol{x}}_{i}$:
\begin{equation}
  \begin{aligned}
  &\hat{\boldsymbol{x}}_{i, d}= \begin{cases}\boldsymbol{x}_{j, d} \text {, if } \operatorname{randb}(d) \leqslant C R \\ \boldsymbol{x}_{i, d} \text {, otherwise } \end{cases} \\
  &\text {s.t.}~~i \ne j, \bblue{d \in \{ 1, 2, \ldots, D \}} \text {, }
  \end{aligned}
\end{equation}
where $\operatorname{randb}(d)$ is the $d$-th evaluation of a uniform random number generator with outcome in $[0, 1]$, and $CR$ is the crossover constant in $[0, 1]$ that is determined by the user. We re-formulate this process as:
\begin{equation}
 \resizebox{.9\hsize}{!}{$
  \begin{aligned}
    \hat{\boldsymbol{x}}_{i}  &= boldsymbol{x}_{i, 1}\boldsymbol{w}_{i, 1}, \dots, \boldsymbol{x}_{i, D}\boldsymbol{w}_{i, D}] + [\boldsymbol{x}_{j, 1}\boldsymbol{w}_{j, 1}, \dots, \boldsymbol{x}_{j, D}\boldsymbol{w}_{j, D}] \\
                              &= \boldsymbol{x}_{1} \odot \boldsymbol{0} + \dots \boldsymbol{x}_{i} \odot \boldsymbol{w}_{i} + \dots \boldsymbol{x}_{j} \odot \boldsymbol{w}_{j} + \dots \boldsymbol{x}_{L} \odot \boldsymbol{0} \\
                              &= \boldsymbol{x}_{1} \boldsymbol{0} + \dots \boldsymbol{x}_{i} \boldsymbol{W}^{cr}_{i} + \dots \boldsymbol{x}_{j} \boldsymbol{W}^{cr}_{j} + \dots \boldsymbol{x}_{L} \boldsymbol{0} \\
                              &= \sum_{l=1}^{L} \left( \boldsymbol{x}_{l} \boldsymbol{W}^{cr}_{l} \right), \\
    & ~\text {s.t.}~~\boldsymbol{w}_{i} + \boldsymbol{w}_{j} = \boldsymbol{1} ~ \text {, }\\
    & ~~~~~~~\boldsymbol{w}_{i, d} \in [0, 1], ~\boldsymbol{w}_{j, d} \in [0, 1], ~\bblue{d \in \{ 1, 2, \ldots, D \}}~ \text {, }
  \end{aligned}
  $}
  \label{eq:crossover}
\end{equation}
where $\boldsymbol{w}_{i}$ and $\boldsymbol{w}_{j}$ are vectors filled with zeros or ones, indicating the feature selections of $\boldsymbol{x}_{i}$ and $\boldsymbol{x}_{j}$, while $\boldsymbol{W}^{cr}_{i}$ and $\boldsymbol{W}^{cr}_{j}$ are corresponding diagonal matrix representations. $\odot$ means the point-wise multiplication operation for each position. $\boldsymbol{0}$ represents that corresponding individual has no contribution, \ie, $\boldsymbol{W}^{cr}_{l} (l \ne i,j)$ \emph{fulls} of zeros. As can be seen above, crossover operator is actually a sparse global feature interaction process.

\emph{For the MSA module of Transformer}, each patch embedding interacts with all embeddings in dense communications. Without loss of generality, $\boldsymbol{x}_{i}$ interacts with the whole population $\boldsymbol{X}$ as follows:
\begin{equation}
  \begin{aligned}
    \hat{\boldsymbol{x}}_{i}  &= \bigoplus_{h=1}^{H} \hat{\boldsymbol{x}}_{i, h} \\
                              &= \bigoplus_{h=1}^{H} \sum_{l=1}^{L} \boldsymbol{A}_{l,h}\boldsymbol{V}_{l,h} \\
                              &= \bigoplus_{h=1}^{H} \sum_{l=1}^{L} \boldsymbol{A}_{l,h}\boldsymbol{x}_{l}\boldsymbol{W}_{h}^{V} \\
                              &= \bigoplus_{h=1}^{H} \sum_{l=1}^{L} \left( \boldsymbol{A}_{l,h} \boldsymbol{x}_{l} \boldsymbol{W}_{h}^{V} \right) \\
                              & = \sum_{l=1}^{L} \left( \boldsymbol{x}_{l} \bigoplus_{h=1}^{H} \left( \boldsymbol{A}_{l,h} \boldsymbol{W}_{h}^{V} \right) \right) \\
                              &= \sum_{l=1}^{L} \left( \boldsymbol{x}_{l} \left( \boldsymbol{A}_{l}\boldsymbol{W}^{V} \right) \right),
  \end{aligned}
  \label{eq:msa}
\end{equation}
where $\boldsymbol{A}_{l, h}$ (\bblue{l $\in$ $\{$1,2,$\cdots$,L$\}$}) is the attention weight of $h$-th head from embedding token $\boldsymbol{x}_{l}$ to $\boldsymbol{x}_{i}$, which is calculated between the query value of $\boldsymbol{x}_{i, h}$ and the key value of $\boldsymbol{x}_{l, h}$ of $h$-th head followed with a $\operatorname{Softmax}(\cdot)$ postprocessing; $\boldsymbol{V}_{l, h}$ (\bblue{l $\in$ $\{$1,2,$\cdots$,L$\}$}) is the projected Value feature for $\boldsymbol{x}_{l}$ with corresponding weights $\boldsymbol{W}_{h}^{V}$; $\hat{\boldsymbol{x}}_{i, h}$ is the sum of all weighted $\boldsymbol{V}_{l, h}$ (\bblue{l $\in$ $\{$1,2,$\cdots$,L$\}$}) by $\boldsymbol{A}_{l, h}$ (\bblue{l $\in$ $\{$1,2,$\cdots$,L$\}$}), \ie, $\hat{\boldsymbol{x}}_{i, h}=\sum_{l=1}^{L} \boldsymbol{A}_{l,h}\boldsymbol{V}_{l,h}$, (\cf, Eq.~\ref{eq:tr_msa} in Section~\ref{section:pre}) for more details. $\boldsymbol{W}^{V}$ is the parameter matrix for the value projection and $\oplus$ means the concatenation operation. By comparing \bblue{Equation~\ref{eq:crossover}} with \bblue{Equation~\ref{eq:msa}}, we find that both above components have the same formula representation, and the crossover operation is a sparse global interaction while densely-modeling MSA has more complex computing and modeling capabilities.

\item[\ding{80}] \textbf{Mutation Operator} \vs \textbf{FFN Module}. \\
\emph{For the mutation operator in EA}, it brings random evolutions into the population by stochastically changing specific features of individuals. Specifically, an individual $\boldsymbol{x}_{i}$ in the population goes through Mutation operation to form the new individual $\hat{\boldsymbol{x}}_{i}$, formulated as follows:
\begin{equation}
  \begin{aligned}
  &\hat{\boldsymbol{x}}_{i, d}= \begin{cases}\operatorname{rand}(v_{d}^L, v_{d}^H)\cdot x_{i, d} \text {, if } \operatorname{randb}(d) \leqslant M U \\ 
    1\cdot x_{i, d} \text {, otherwise } \end{cases} \\
  &\text {s.t.}~~\bblue{d \in \{ 1, 2, \ldots, D \}} \text {, }
  \end{aligned}
\end{equation}
where $\operatorname{randb}(d)$ is the $d$-th evaluation of a uniform random number generator with outcome in $[0, 1]$, and $MU$ is the mutation constant in $[0, 1]$ that user determines. $v_{j}^L$ and $v_{j}^H$ are lower and upper scale bounds of the $j$-th feature relative to $x_{i, d}$. Similarly, we re-formulate this process as:
\begin{equation}
  \begin{aligned}
    \hat{\boldsymbol{x}}_{i}  &= \boldsymbol{x}_{i} \odot \boldsymbol{w}_{i} \\
                              &= \boldsymbol{x}_{i} \boldsymbol{W}^{mu}_{i},
  \end{aligned}
  \label{eq:mutation}
\end{equation}
where $\boldsymbol{w}_{i}$ is a randomly generated vector that represents weights of each feature value, while $\boldsymbol{W}^{mu}_{i}$ is the corresponding diagonal matrix representation; $\odot$ means the point-wise multiplication operation for each position.

\emph{For the FFN module in Transformer}, each patch embedding carries on directional feature transformation through cascaded linear layers (\cf, Equation~\ref{eq:tr_ffn}). Getting rid of complex nonlinear transformations, we only take one linear layer as an example:
\begin{equation}
  \begin{aligned}
    \hat{\boldsymbol{x}}_{i}  &= \boldsymbol{x}_{i} \boldsymbol{W}^{FFN}, \\
  \end{aligned}
  \label{eq:ffn}
\end{equation}
where $\boldsymbol{W}^{FFN}$ is the weight of the linear layer, and it is applied to each embedding separately and identically. 

By analyzing the calculation process of Equation~\ref{eq:mutation} and Equation~\ref{eq:ffn}, Mutation and FFN operations share a unified form of matrix multiplication, so they are supposed to own a consistent function essentially. 
\textbf{\textit{Besides}}, at the microcosmic level, the weight of FFN change dynamically during the training process, so the output of the individual differs among different iterations (similar to the random process of mutation). At the macroscopic objective of the algorithm, the mutation in EA is optimized into one potential direction under the constraint of the objective function (statistically speaking, only partial mutation individuals are retained, that is, the mutation also has a determinate meaning in the whole training process). In comparison, the trained FFN can be regarded as a directional mutation under the constraint of loss functions. 
\textbf{\textit{Finally}}, note that we are only discussing the comparison with the mutation on one linear layer of FFN, and $\boldsymbol{W}^{FFN}$ is more expressive than diagonal $\boldsymbol{W}^{mu}_{i}$ in fact because it contains cascaded linear layers and the non-linear ReLU activation is interspersed between adjacent linear layers, as depicted in Equation~\ref{eq:tr_ffn}.

\item[\ding{80}] \textbf{Population Succession} \vs \textbf{RC Operation}. \\
In the evolution of the biological population, individuals at the current iteration have a certain probability of inheriting to the next iteration, where a partial population of the current iteration will be combined with the selected individuals. Similarly, the above pattern is expressed by Transformer structure in the form of Residual Connection (RC), \ie, patch embeddings of the previous layer are directly mapped to the next layer. Specifically, partial-selection can be viewed as a dropout technique in Transformer, while population succession can be formulated as a concatenation operation that has a consistent mathematical expression with residual concatenation, whereas addition operation can be regarded as a particular case of the concatenation operation that shares some partial weights.

\item[\ding{80}] \textbf{Best Individual} \vs \textbf{Task-Related Token}. \\
Generally speaking, the Transformer-based model chooses an enhanced task-related token (\eg, classification token) that combines information of all patch embeddings as the output feature, while the EA-based method chooses the individual with the best fitness score among the population as the output.

\begin{figure*}[thp]
  \centering
  \includegraphics[width=1.0\linewidth]{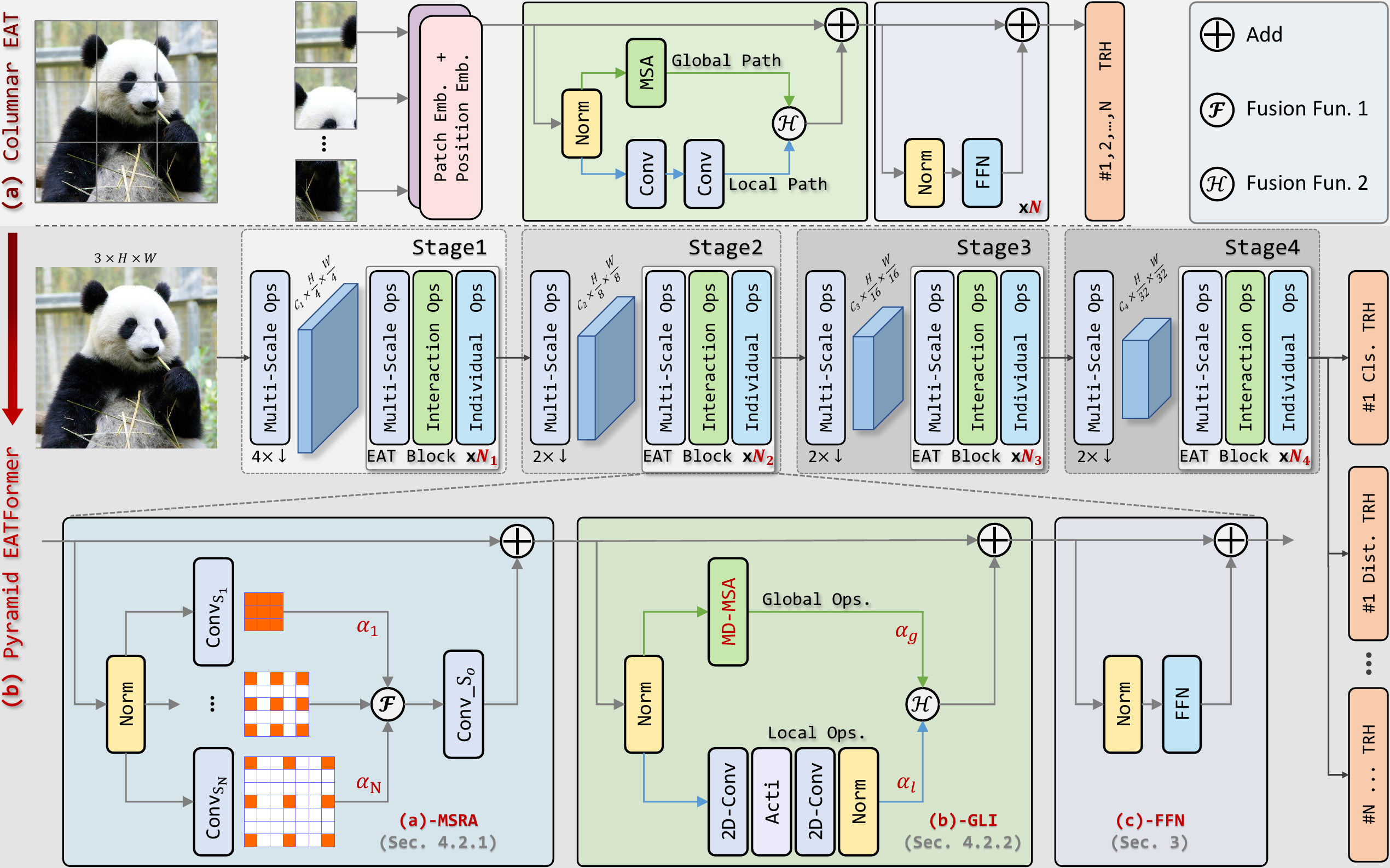}
  \caption{Structure of EA-inspired columnar EAT model~\cite{eat} and improved pyramid EATFormer. \textcolor[RGB]{125,0,0}{\textbf{(a)}} The \emph{top} part shows the architecture of the previous EAT model, where the basic block consists of parallel global and local paths as well as an FFN module. \textcolor[RGB]{192,0,0}{\textbf{(b)}} The \emph{middle} part illustrates overall architecture of EATFormer that contains four stages with $i$-th stage consisting of \textcolor[RGB]{192,0,0}{\textbf{\emph{N}$_i$}} basic EAT blocks; The \emph{bottom} part illustrates the structure of serial modules in EAT block, \ie, MSRA (\cf, Section~\ref{section:msra}), GLI (\cf, Section~\ref{section:gli}), and FFN (\cf, Section~\ref{section:pre}) from left to right, and a MD-MSA is proposed to effectively improve the model performance; The right part shows the designed Task-Related Head module docked with transformer backbone for specific tasks.}
  \label{fig:net}
\end{figure*}

\item[\ding{80}] \textbf{Necessity of Modules in Transformer.} \\
As described in the work~\cite{s16}, the absence of the crossover operator or mutation operator will significantly damage the model's performance. Similarly, Dong~\etal~\cite{attn_notattention} explore the effect of MLP in the Transformer and find that MLP stops the output from degeneration, and removing MSA in Transformer would also significantly damage the effectiveness of the model. Thus we can conclude that global information interaction and individual evolution are necessary for Transformer, just like the global crossover and individual mutation in EA.
\end{description}

\subsection{Short Description of Previous Columnar EAT} \label{section:eat}
We explore the relationship among operators in naive EA and modules in naive Transformer in the previous NeurIPS'21 conference\cite{eat} and analogically improve a columnar EAT based on ViT model. Figure~\ref{fig:net}-\textcolor[RGB]{125,0,0}{\textbf{(a)}} shows the structure of EAT model that is stacked of with \textcolor[RGB]{192,0,0}{\textbf{\emph{N}}} improved Transformer blocks inspired by local population concept in some EA works~\cite{lea1,lea2,mea1}, where a local path is introduced in parallel with global MSA operation. Also, this work designs a Task-Related Head to deal with various tasks more flexibly, \ie, classification and distillation.

However, the columnar structure is naturally inadequate for downstream dense prediction tasks, and it is inferior in terms of accuracy compared with contemporaneous works~\cite{attn_pvt,attn_swin}, which limits the usefulness of the model in some scenarios. To address the above weaknesses, this paper further explores analogies between EA and Transformer and improves the previous work to a pyramid EATFormer, consisting of the newly designed EAT block inspired by the effective EA variants. 

\subsection{Methodology of Pyramid EATFormer Architecture} \label{section:eatformer}
Architecture of the improved EATFormer is illustrated in Figure~\ref{fig:net}-\textcolor[RGB]{192,0,0}{\textbf{(b)}}, which contains four stages of different resolutions following PVT~\cite{attn_pvt}. Specifically, the model is made up of EAT blocks that contains three mixed-paradigm $y = f(x) + x$ residuals: (a) Multi-Scale Region Aggregation (MSRA), (b) Global and Local Interaction (GLI), and (c) Feed-Forward Network (FFN) modules, and the down-sampling procedure between two stages is realized by MSRA with stride greater than 1. Besides, we propose a novel Modulated Deformable MSA (MD-MSA) to advance global modeling and a Task-Related Head (TRH) to complete different tasks more elegantly and flexibly. 

\subsubsection{Multi-Scale Region Aggregation} \label{section:msra}
Inspired by some multi-population-based EA methods~\cite{lpea2,lpea4} that would adopt different searching regions for obtaining a better model performance, we analogically extend this concept to multiple sets of spatial positions for the 2D image and design a novel \emph{Multi-Scale Region Aggregation} (MSRA) module for the studied vision transformer. As shown in Figure~\ref{fig:net}.(a), MSRA contains $N$ local convolution operations (\ie, ${\rm Conv_{{S}_{n}}}, 1 \leq n \leq N $) with different strides to aggregate information from different receptive fields, which simultaneously \emph{play} the role of providing inductive bias without extra position embedding procedures. Specifically, the $n$-th dilation operation $o_{n}$ that transforms input feature map $x$ can be formulated as:
\begin{equation}
  \begin{aligned}
    &o_{n}(\boldsymbol{x}) = \text{Conv}_{{S}_{n}}(\text{Norm}(\boldsymbol{x})) \\
    &\text{s.t.}~~\bblue{n \in \{ 1, 2, \ldots, N \}} \text {, }
  \end{aligned}
  \label{eq:msra_path}
\end{equation}
\emph{Weighted Operation Mixing} (WOM) mechanism is further proposed to mix all operations by a softmax function over a set of learnable weights $\alpha_{1}, \dots, \alpha_{N}$, and the intermediate representation $\boldsymbol{x}_o$ is calculated by the mixing function $\mathcal{F}$ as follows:
\begin{equation}
  \begin{aligned}
    \boldsymbol{x}_{o} = \sum_{n=1}^{N} \frac{\exp \left(\alpha_{n}\right)}{\sum_{n^{\prime}=1}^{N} \exp \left(\alpha_{n^{\prime}}\right)} o_{n}(\boldsymbol{x}), \\
  \end{aligned}
  \label{eq:msra}
\end{equation}
where $\mathcal{F}$ in the above formula is the addition function, and other fusion functions like concatenation are also available for a better effect at the cost of more parameters. The paper chooses the addition function acquiescently. Then, a convolution layer ${\rm Conv_{{S}_{o}}}$ maps $\boldsymbol{x}_{o}$ to the same number of channels as the input $\boldsymbol{x}$, and the final output of the module is obtained after a residual connection. Also, the MSRA module serves as the \emph{model stem} and \emph{Patch Embedding} that makes the EATFormer more uniform and elegant. Note that this paper does not use any form of position embedding since CNN-based MSRA can provide a natural inductive bias for the next GLI module.

\subsubsection{Global and Local Interaction} \label{section:gli}
Motivated by EA variants~\cite{mea1,mea2,mea3} that introduce local search procedures besides conventional global search for converging higher-quality solutions faster and effectively (\cf, Figure~\ref{fig:motivation}-(c) for a better intuitive explanation), we improve a MSA-based global module to a novel \emph{Global and Local Interaction} (GLI) module. As shown in Figure~\ref{fig:net}.(b), GLI contains an extra local path in parallel with the global path, where the former aims to mine more discriminative locality-relevant information like the above-mentioned local population idea, while the latter is retained to model global information. Specifically, the input features are divided into global features (marked green) and local features (marked blue) at the channel level with ratio $p$, which are then fed into global and local paths to conduct feature interactions, respectively. Note that we also apply the proposed \emph{Weighted Operation Mixing} mechanism in~\ref{section:msra} to balance two branches, \ie, global weight $\alpha_{\textit{g}}$ and local weight $\alpha_{\textit{l}}$. The outputs of the two paths recover the original data dimension by concatenation operation $\mathcal{H}$. Thus the improved module is very flexible and can be viewed as a plug-and-play module for the current transformer structure. In detail, the local operation can be traditional convolution layer or other improved modules, \eg, DCN~\cite{dcn1,dcn2}, local MSA, \emph{etc}, while global operation can be MSA~\cite{attn,attn_vit}, D-MSA~\cite{attn_dpt}, Performer~\cite{attn_improve3}, \etc

In this paper, we choose naive convolution with MSA modules as basic compositions of GLI, and it owns $O(1)$ maximum path length between any two positions for keeping global modeling capability besides enhancing locality, as shown in Table~\ref{tab:params_flops}. Therefore, the proposed structure maintains the same parallelism and efficiency as the original vision transformer. Also, the selection of feature separation ratio $p$ is crucial to the effect and efficiency of the model because different ratios bring different parameters, FLOPs, and precision of the model. In detail, the local path contains a group of point-wise and $k \times k$ depth-wise convolutions. Assume that the feature map in $\mathbb{R}^{C \times H \times W}=\mathbb{R}^{C \times L}$, and both paths have $C_g = p \times C$ and $C_l = C - C_g$ channels, respectively. Here we present an analysis process about the number of parameters and computation of the improved GLI module as follows:

\noindent \textbf{\emph{1) Overall Params}} equals $4(C_{g}+1)C_{g} + (k^{2}+1)C_{l} + (C_{l}+1)C_{l}$ according to Table~\ref{tab:params_flops}, and it is factorized based on $C_{l} = C - C_{g}$:
\begin{equation}
  \begin{aligned}
    Params =  &5{C_{g}}^2 + (2-2C-K^{2})C_{g} + \\
              &(k^{2} + 2 + C)C.
   \end{aligned}
   \label{eq:params}
\end{equation}
Applying the minimum value formula of a quadratic function, Equation~\ref{eq:params} obtains the minimum value when $C_{g}^{min_p} = 0.2C + 0.1(k^{2} - 2)$. Given that the channel number are integers and latter term can be ignored, we obtains $C_{g}^{min_p} = 0.2C$, \ie, $p^{min_p}$ equals 0.2.

\noindent \textbf{\emph{2) Overall FLOPs}} equals $8C_{g}^2L+4C_{g}L^2+3L^2 + (2k^{2})LC_{l} + 2C_{l}LC_{l}$ according to Table~\ref{tab:params_flops}, and it is factorized based on $C_{l} = C - C_{g}$:
\begin{equation}
  \begin{aligned}
    FLOPs = &10L{C_{g}}^2 + (4L^{2} - 2k^{2}L - 4LC)C_{g} + \\
            &(3L+ 2k^{2}C + 2C^{2})L.
  \end{aligned}
  \label{eq:flops}
\end{equation}
Applying the minimum value formula of a quadratic function, Equation~\ref{eq:flops} obtains the minimum value when $C_{g}^{min_f} = 0.2C + 0.1(k^{2} - 2L)$. Also, ignoring the latter term, we obtains $C_{g}^{min_f} = 0.2C$ that follows the same trend with $C_{g}^{min_p}$. Therefore, we can draw two conclusions: \ding{172} The parameters and calculations of GLI are much lower than single-path MSA ($p$ < 1), and the minimum value can be obtained when using both paths ($p$ > 0); \ding{173} According to Equation~\ref{eq:params} and Equation~\ref{eq:flops}, there is not much difference about the total parameters and calculations when $p$ lies in the range [0, 0.5], so $p$ is set 0.5 for all layers in this paper for simplicity and efficiency. Also, experiments in Section~\ref{explanation:alpha} demonstrate that $p=0.5$ is the most economical and efficient option. Note that the number of convolution parameters and computation of the current local path are smaller compared with the global path, while the stronger local structure will make ratio $p$ change larger and this paper will not elaborate on the details.

Furthermore, we advance the global path by designing a Modulated Deformable MSA (MD-MSA in Section~\ref{section:mdmsa}) module, which improves the model performance with negligible parameters and GFLOPs increasing, and a comparison study to explore combinations of different operations is further conducted in the experimental section. 

\begin{table}[!t]
  \caption{Properties of convolution and MSA layers with Parameters (Params), floating point operations (FLOPs), and Maximum Path Length (MPL). Assume that the input and output feature maps in $\mathbb{R}^{C \times H \times W}$, $L = H \times W$, $H=W$, $k$ and $G$ are kernel size and group number for convolution layers.}
  \centering
  \begin{tabular}{c|c|c|c}
      \toprule
      Type  & Params          & FLOPs               & MPL       \\
      \midrule
      MSA   & $4(C+1)C$       & $8C^2L+4CL^2+3L^2$  & $O(1)$    \\
      Conv  & $(Ck^{2}/G+1)C$ & $(2Ck^{2}/G)LC$     & $O(H/k)$  \\
      \bottomrule
  \end{tabular}
  \label{tab:params_flops}
\end{table}

\begin{figure}[htp]
  \centering
  \includegraphics[width=1.0\linewidth]{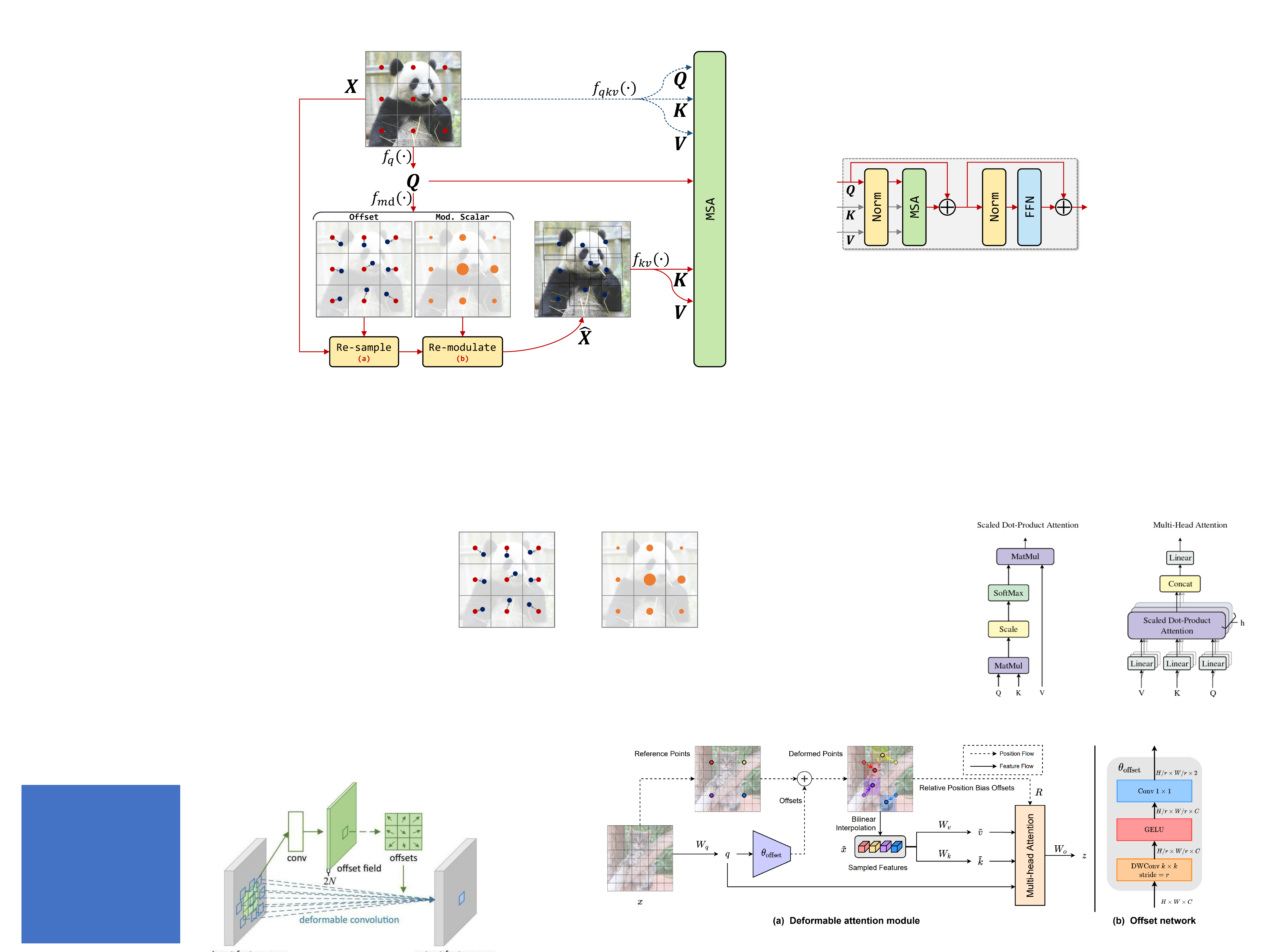}
  \caption{Structure of the proposed MD-MSA.}
  \label{fig:mdmsa}
\end{figure}

\subsubsection{Modulated Deformable MSA} \label{section:mdmsa}
Inspired by the irregular spatial distribution among real individuals that are not as horizontal and vertical as the image, we \emph{improve} a novel Modulated Deformable MSA (MD-MSA) module that considers position fine-tuning and re-weighting of each spatial patch. As shown in Figure~\ref{fig:mdmsa}, the blue dotted line represents naive MSA procedure that $\boldsymbol{QKV}$ features are obtained by the input feature map $\boldsymbol{X}$ from function $f_{qkv}(\cdot)$, \ie, $\boldsymbol{QKV}=f_{qkv}(\boldsymbol{X})$ and $f_{qkv} = f_{q} \oplus f_{k} \oplus f_{v}$ ($\oplus$ denotes concatenation operation), while the red solid line shows the procedure of MD-MSA. And the main difference between the proposed MD-MSA and original MSA lies in the query-aware access of fine-tuned feature map $\boldsymbol{\hat{X}}$ to extract $\boldsymbol{KV}$ features further. Specifically, given the input feature map $\boldsymbol{X}$ with $L$ positions, $\boldsymbol{Q}$ is obtained by function $f_{q}$, \ie, $\boldsymbol{Q}=f_{q}(\boldsymbol{X})$, which is then used to predict deformable offset $\Delta l$ and modulation scalar $\Delta m$ for all positions:
\begin{equation}
  \begin{aligned}
    \Delta l, \Delta m = f_{md}(\boldsymbol{Q}).
  \end{aligned}
  \label{eq:mdmsa}
\end{equation}
For the $l$-th position, the re-sampled and re-weighted feature $\boldsymbol{\hat{X}}_l$ is calculated by:
\begin{equation}
  \begin{aligned}
    \boldsymbol{\hat{X}}_l = \mathcal{S}(\boldsymbol{X}_l, \Delta l) \cdot \Delta m,
  \end{aligned}
  \label{eq:mdmsa1}
\end{equation}
where $\Delta l$ is the relative coordinate with an unconstrained range for the $l$-th position, while $\Delta m$ lies in the range [0, 1], and $\mathcal{S}$ represents the bilinear interpolation function. Then $\boldsymbol{KV}$ is obtained with the new feature map $\boldsymbol{\hat{X}}$, \ie, $\boldsymbol{KV}=f_{kv}(\boldsymbol{\hat{X}})$. It is worth mentioning that the main difference between MD-MSA and recent similar work~\cite{attn_DAT} lies in the modulation operation, where MD-MSA could apply appropriate attention to different position features to obtain better results. Also, any form of position embedding is \emph{not used} since it makes no contribution to results, and detailed comparative experiments can be viewed in Section~\ref{ablation:eatformer}.

\begin{figure}[!t]
  \centering
  \includegraphics[width=0.8\linewidth]{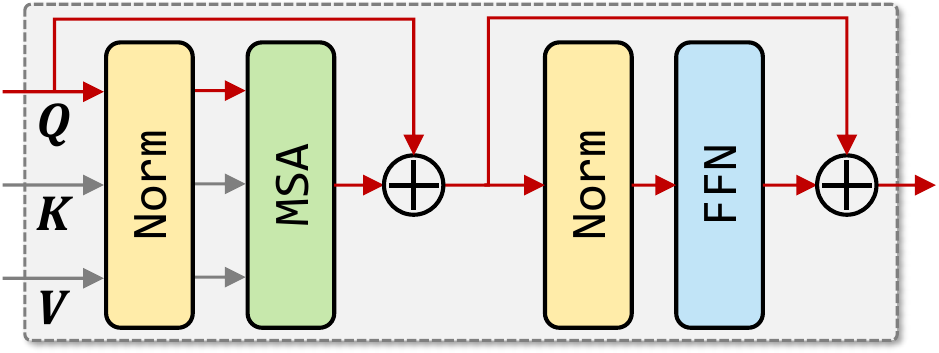}
  \caption{Structure of the proposed Task-Related Head.}
  \label{fig:trh}
\end{figure}

\begin{table*}[htp]
  \centering
  \caption{Overall analogical correlations between EA and EATFormer.}
  \label{table:analogous_table}
  \renewcommand{\arraystretch}{0.88}
  \begin{tabular}{p{0.3cm} | p{5cm}<{\raggedleft} | p{5.6cm}<{\raggedright} p{2.0cm}<{\raggedright}}
  \toprule
  & EA & EATFormer \\
  \midrule
  \multirow{6}{*}{\rotatebox{90}{Basics}} & Population Size                   & Patch Number & \multirow{6}{*}{\rotatebox{0}{(Section~\ref{section:intro})}} \\
  & (Discrete) Individual             & (Continuous) Patch Token & \\
  & (Sparse) Crossover Operation      & (Dense) Global MSA module & \\
  & (Sparse) Mutation Operation       & (Dense) Individual FFN module & \\
  & (Partial) Population Succession   & (Integral) Residual Connection & \\
  & (Global) Best Individual          & (Aggregated) Task-Related Token & \\
  \midrule
  \multirow{6}{*}{\rotatebox{90}{Improvements}} & Multi-Scale Population            & Multi-Scale Region Aggregation module & (Section~\ref{section:msra}) \\ 
  & Global and Local Population       & Global and Local Interaction module   & (Section~\ref{section:gli}) \\
  & Self-Adapting Parameters          & Weighted Operation Mixing             & (Section~\ref{section:gli}) \\
  & Irregular Population Distribution & Modulated Deformable MSA              & (Section~\ref{section:mdmsa}) \\
  & Multi-Objective EA                & Task-Related Head                     & (Section~\ref{section:trh}) \\
  & Dynamic Population                & Pyramid Architecture                  & (Figure~\ref{fig:net}) \\
  \bottomrule
  \end{tabular}
\end{table*}

\subsubsection{Task-Related Head} \label{section:trh}
Current transformer-based vision models would initialize different tokens for different tasks~\cite{attn_deit} or use the pooling operation to obtain global representation~\cite{attn_swin}. However, both manners are potentially incompatible: the former treats the task token and image patches coequally as unreasonable and clumsy, because the task token and image tokens have different feature distributions while additional computation is required from $O(n^2)$ to $O((n+1)^2)$; the latter uses only one pooling result for multiple tasks that is also inappropriate and harmful for losing wealthy information. Inspired by multi-objective EAs~\cite{moea1,moea2,moea3} that find a set of solutions for different targets, we design a \emph{Task-Related Head} (TRH) docked with transformer backbone to obtain the corresponding task output through the final features. As shown in Figure~\ref{fig:trh}, we employ a cross-attention paradigm to implement this module: $K$ and $V$ (gray lines) are output features extracted by the transformer backbone, while $Q$ (red line) is the task-related token to integrate global information. Note that this design is more effective and flexible for different tasks learning simultaneously while consuming a negligible computation amount compared to the backbone, and more analytical experiments can be viewed in the following Section~\ref{ablation:trh}. For a more fair comparison, TRH presented in the former conference version~\cite{eat} is not used by default because this plug-and-play module can easily be added to other methods, and we will conduct an ablation experiment in Section~\ref{ablation:trh} to verify the validity of TRH.

\subsubsection{Overall Congruent Relationships}
To more clearly show the design inspirations of different modules, we summarize the analogies between the improved EATFormer research and homologous concepts (ideas) from EA variants in Table~\ref{table:analogous_table}.

\vspace{-1.0em}
\subsection{EATFormer Variants}
In the former conference version~\cite{eat}, we improve the columnar ViT by introducing a local path in parallel with global MSA operation, denoted as EAT-Ti, EAT-S, and EAT-B in the top part of Table~\ref{tab:variants}. In this paper, we extend the columnar structure to a pyramid architecture and carefully re-design a novel EATFormer model, which has a series of scales for different practical applications, and these variants can be viewed in the bottom part of Table~\ref{tab:variants}. Except for the depth and dimension of the model, other parameters remain consistent for all models: the head dimension of MSA is 32; window size is set to 7; kernel size of all convolution is $3 \times 3$; dilations of the MSRA module for four stages are [1], [1], [1,2,3], and [1,2], respectively; low-level stage1-2 only use local path while high-level stage3-4 employ hybrid GLI module for efficiency. More detailed structures and implementations can be viewed in the attached source code.

\begin{table*}[htp]
  \centering
  \caption{Detailed settings of different EATFormer variants. Top part shows previous columnar EAT models~\cite{eat}.}
  \label{tab:variants}
  \renewcommand{\arraystretch}{0.88}
  \setlength\tabcolsep{3.0pt}
  \begin{tabular}{p{0.2cm} | p{3.0cm}<{\raggedright} | p{2.0cm}<{\centering} | p{3.0cm}<{\centering} | p{1.7cm}<{\centering} | p{1.7cm}<{\centering} | p{2.3cm}<{\centering} | p{1.2cm}<{\centering}}
  \toprule
  & \pzo\pzo\pzo\pzo\pzo\pzo Network & Depth & Dimension & Params.~(M) & FLOPs~(G) & Inf. Mem.~(G) & Top-1 \\
  \midrule
  \multirow{3}{*}{\rotatebox{90}{Col.}} & EAT-Ti      & 12                & 192                     & \pzo\pzo5.7 & \pzo1.01  & \pzo2.2 & 72.7 \\
  & EAT-S       & 12                & 384                     & \pzo22.1    & \pzo3.83  & \pzo2.9 & 80.4 \\
  & EAT-B       & 12                & 768                     & 86.6        & 14.83     & \pzo4.5 & 82.0 \\
  \midrule
  \multirow{9}{*}{\rotatebox{90}{Pyramid}} & EATFormer-Mobile  & [ 1, 1, 4, 1 ]    & [ 48, \pzo64, 160, 256 ]    & \pzo\pzo1.8 & \pzo0.36  & \pzo2.2 & 69.4 \\
  & EATFormer-Lite    & [ 1, 2, 6, 1 ]    & [ 64, 128, 192, 256 ]   & \pzo\pzo3.5 & \pzo0.91  & \pzo2.7 & 75.4 \\
  & EATFormer-Tiny    & [ 2, 2, 6, 2 ]    & [ 64, 128, 192, 256 ]   & \pzo\pzo6.1 & \pzo1.41  & \pzo3.1 & 78.4 \\
  & EATFormer-Mini    & [ 2, 3, 8, 2 ]    & [ 64, 128, 256, 320 ]   & \pzo11.1    & \pzo2.29  & \pzo3.6 & 80.9 \\
  & EATFormer-Small   & [ 3, 4, 12, 3 ]   & [ 64, 128, 320, 448 ]   & \pzo24.3    & \pzo4.32  & \pzo4.9 & 83.1 \\
  & EATFormer-Medium  & [ 4, 5, 14, 4 ]   & [ 64, 160, 384, 512 ]   & \pzo39.9    & \pzo7.07  & \pzo6.2 & 83.6 \\
  & EATFormer-Base    & [ 5, 6, 20, 7 ]   & [ 96, 160, 384, 576 ]   & \pzo63.5    & 10.89     & \pzo8.7 & 83.9 \\
  \bottomrule
  \end{tabular}
\end{table*}

\vspace{-1.0em}
\subsection{Further Discussion}
Compared with EAT in the former conference version, the improved EATFormer has better inspirations, finer analogical designs, and more sufficient experiments. And we prove the effectiveness and integrity of the proposed method through a series of following experiments, such as comparison with SOTA methods, downstream task transferring, ablation studies, and explanatory experiments. It is worth noting that the backbone of EATFormer in this paper only contains one unified EAT block, which fully considers three aspects of modeling: \emph{1)} multi-scale information aggregation, \emph{2)} feature interactions among tokens, and \emph{3)} individual enhancement. Also, the architecture recipes of EATFormer variants in this paper are mainly given by our intuition and proved by experiments, but the alterable configure parameters can be used as the search space for NAS that is worth further exploration in our future works, \eg, embedding dimension, dilations of MSRA, kernel size of MSRA, fusion function of MSRA, down-sampling mode of MSRA, separation ratio of GLI, normalization types, window size, operation combinations of GLI, \etc

\section{Experiments} \label{section:exp}
In this section, to evaluate the effectiveness and superiority of our improved EATFormer architecture, we experiment for mainstream vision tasks with models of different volumes as the backbone and orderly conduct down-stream tasks, \ie, \emph{image-level classification} (ImageNet-1K~\cite{imagenet}), \emph{object-level detection and instance segmentation} (COCO 2017~\cite{coco}), and \emph{pixel-level semantic segmentation} (ADE20K~\cite{ade20k}). Massive ablation and explanatory experiments are further conducted to prove the effectiveness of EATFormer and its components.

\subsection{Image Classification}
\subsubsection{Experimental Setting} \label{section:exp_cls}
\textit{All of our EATFormer variants are trained for 300 epochs from scratch \textbf{without} pre-training, extra datasets, pre-trained models, token labeling~\cite{tlt} alike strategy, and exponential moving average.} We employ the same training recipe as Deit~\cite{attn_deit} to all EATFormer variants for fair comparisons with different SOTA methods: AdamW~\cite{adamw} optimizer is used for training with betas and weight decay equaling (0.9, 0.999) and 5$e^{-2}$, respectively; Batch size is set to 2,048, while learning rate is 5$e^{-4}$ by default with a linear increasing compared with batch size divided by 512; Standard cosine learning rate scheduler, data augmentation strategies, warm-up, and stochastic depth are used during the training phase~\cite{attn_deit}. EATFormer is built on PyTorch~\cite{pytorch} and relies on the TIMM interface~\cite{timm}.

\begin{table*}[htp]
  \centering
  \caption{Comparison with SOTA methods on ImageNet-1K. Reported results are from corresponding papers. \protect\sethlcolor{lgray_tab}\hl{Gray} background shows CNN-based and columnar methods, while the proposed \blue{EATFormer variants} are colored in \protect\sethlcolor{lblue_tab}\hl{blue}.}
  \vspace{-0.8em}
  \label{table:imagenet}
  \renewcommand{\arraystretch}{0.6}
  \resizebox{0.92\linewidth}{!}{
      \begin{tabular}{p{3.6cm}<{\raggedright} p{1.2cm}<{\centering} p{1.7cm}<{\centering} p{1.1cm}<{\centering} p{1.1cm}<{\centering} p{1.7cm}<{\centering} p{1.2cm}<{\centering} p{1.7cm}<{\raggedleft}}
      \hline
      \multirow{2}{*}{\pzo\pzo\pzo\pzo\pzo\pzo\pzo\pzo Network} & \multirow{2}{*}{\makecell[c]{Params. $\downarrow$ \\(M)}} & \multirow{2}{*}{\makecell[c]{FLOPs $\downarrow$ \\(G)}} & \multicolumn{2}{c}{\makecell[c]{Images/s $\uparrow$}} & \multirow{2}{*}{\makecell[c]{Resolution}} & \multirow{2}{*}{Top-1} & \multirow{2}{*}{\makecell[c]{Pub.\pzo\pzo}} \\
      \cmidrule(lr){4-5}
      & & & GPU & CPU & & & \\
      \hline
      \rowcolor{lgray_tab}
      MNetV3-Small 0.75x~\cite{mnetv3}      & \pzo2.0   & \pzo0.05  & 9872      & 589.1         & $224^{2}$ & 65.4  & ICCV'19   \\
      \rowcolor{lblue_tab}
      \textbf{EATFormer-Mobile}             & \pzo1.8   & \pzo0.36  & 3926      & 456.3         & $224^{2}$ & 69.4  & -\pzo\pzo\pzo \\
      \hline
      \rowcolor{lgray_tab}
      MobileNetV3 0.75×~\cite{mnetv3}       & \pzo4.0   & \pzo0.16  & 5585      & 315.4         & $224^{2}$ & 73.3  & ICCV'19   \\
      PVTv2-B0~\cite{attn_pvt}              & \pzo3.6   & \pzo0.57  & 1711      & 104.2         & $224^{2}$ & 70.5  & ICCV'21   \\
      XCiT-N12~\cite{attn_xcit}             & \pzo3.1   & \pzo0.56  & 2736      & 290.9         & $224^{2}$ & 69.9  & NeurIPS'21\\
      VAN-Tiny~\cite{attn_van}              & \pzo4.1   & \pzo0.87  & 1706      & 107.9         & $224^{2}$ & 75.4  & arXiv'22  \\
      \rowcolor{lblue_tab}
      \textbf{EATFormer-Lite}               & \pzo3.5   & \pzo0.91  & 2168      & 246.3         & $224^{2}$ & 75.4  & -\pzo\pzo\pzo \\
      \hline
      \rowcolor{lgray_tab}
      DeiT-Ti~\cite{attn_deit}              & \pzo5.7   & \pzo1.25  & 2342      & 417.7         & $224^{2}$ & 72.2  & ICML'21   \\
      \rowcolor{lgray_tab}
      \textbf{EAT-Ti}~\cite{eat}            & \pzo5.8   & \pzo1.01  & 2356      & 436.8         & $224^{2}$ & 72.7  & NeurIPS'21\\
      \rowcolor{lgray_tab}
      EfficientNet-B0~\cite{efficientnet}   & \pzo5.3   & \pzo0.39  & 2835      & 225.1         & $224^{2}$ & 77.1  & ICML'19   \\
      CoaT-Lite Tiny~\cite{attn_coat}       & \pzo5.7   & \pzo1.59  & 1055      & 143.5         & $224^{2}$ & 77.5  & ICCV'21   \\
      ViTAE-6M~\cite{attn_vitae}            & \pzo6.6   & \pzo2.16  & \pzo921   & 152.6         & $224^{2}$ & 77.9  & NeurIPS'21\\
      XCiT-T12~\cite{attn_xcit}             & \pzo6.7   & \pzo1.25  & 1750      & 259.5         & $224^{2}$ & 77.1  & NeurIPS'21\\
      MPViT-T~\cite{attn_mpvit}             & \pzo5.8   & \pzo1.65  & \pzo755   & 125.9         & $224^{2}$ & 78.2  & CVPR'22   \\
      \rowcolor{lblue_tab}
      \textbf{EATFormer-Tiny}               & \pzo6.1   & \pzo1.41  & 1549      & 167.5         & $224^{2}$ & 78.4  & -\pzo\pzo\pzo \\
      \rowcolor{lblue_tab}
      \textbf{EATFormer-Tiny-384}           & \pzo6.1   & \pzo4.23  & \pzo536   & \pzo56.9      & $384^{2}$ & 80.1  & -\pzo\pzo\pzo \\
      \hline
      \rowcolor{lgray_tab}
      EfficientNet-B2~\cite{efficientnet}   & \pzo9.1   & \pzo0.88  & 1440      & 143.4         & $256^{2}$ & 80.1  & ICML'19   \\
      PVTv2-B1~\cite{attn_pvt}              & 14.0      & \pzo2.12  & 1006      & \pzo79.2      & $224^{2}$ & 78.7  & ICCV'21   \\
      ViTAE-13M~\cite{attn_vitae}           & 10.8      & \pzo3.05  & \pzo698   & 114.3         & $224^{2}$ & 81.0  & NeurIPS'21\\
      XCiT-T24~\cite{attn_xcit}             & 12.1      & \pzo2.35  & \pzo933   & 146.6         & $224^{2}$ & 79.4  & NeurIPS'21\\
      CoaT-Lite Mini~\cite{attn_coat}       & 11.0      & \pzo1.99  & \pzo968   & 164.8         & $224^{2}$ & 79.1  & ICCV'21   \\
      PoolFormer-S12~\cite{attn_metaformer} & 11.9      & \pzo1.82  & 1858      & 218.7         & $224^{2}$ & 77.2  & CVPR'22   \\
      MPViT-XS~\cite{attn_mpvit}            & 10.5      & \pzo2.97  & \pzo612   & \pzo95.8      & $224^{2}$ & 80.9  & CVPR'22   \\
      VAN-Small~\cite{attn_van}             & 13.8      & \pzo2.50  & \pzo992   & \pzo95.2      & $224^{2}$ & 81.1  & arXiv'22  \\
      \rowcolor{lblue_tab}
      \textbf{EATFormer-Mini}               & 11.1      & \pzo2.29  & 1055      & 122.1         & $224^{2}$ & 80.9  & -\pzo\pzo\pzo \\
      \hline
      \rowcolor{lgray_tab}
      DeiT-S~\cite{attn_deit}               & 22.0      & \pzo4.60  & \pzo937   & 163.5         & $224^{2}$ & 79.8  & ICML'21   \\
      \rowcolor{lgray_tab}
      \textbf{EAT-S}~\cite{eat}             & 22.1      & \pzo3.83  & \pzo964   & 175.6         & $224^{2}$ & 80.4  & NeurIPS'21\\
      \rowcolor{lgray_tab}
      ResNet-50~\cite{resnet,timm_resnet}   & 25.5      & \pzo4.11  & 1192      & 123.8         & $224^{2}$ & 80.4  & CVPR'16   \\
      \rowcolor{lgray_tab}
      EfficientNet-B4~\cite{efficientnet}   & 19.3      & \pzo3.13  & \pzo495   & \pzo38.2      & $320^{2}$ & 82.9  & ICML'19   \\
      \rowcolor{lgray_tab}
      CoaT-Lite Small~\cite{attn_coat}      & 19.8      & \pzo3.96  & \pzo542   & \pzo93.8      & $224^{2}$ & 81.9  & ICCV'21   \\
      PVTv2-B2~\cite{attn_pvt}              & 25.3      & \pzo4.04  & \pzo585   & \pzo36.5      & $224^{2}$ & 82.0  & ICCV'21   \\
      Swin-T~\cite{attn_swin}               & 28.2      & \pzo4.50  & \pzo664   & \pzo88.0      & $224^{2}$ & 81.3  & ICCV'21   \\
      XCiT-S12~\cite{attn_xcit}             & 26.2      & 18.92     & \pzo187   & \pzo30.1      & $224^{2}$ & 82.0  & NeurIPS'21\\
      ViTAE-S~\cite{attn_vitae}             & 24.0      & \pzo6.20  & \pzo399   & \pzo69.9      & $224^{2}$ & 82.0  & NeurIPS'21\\
      UniFormer-S~\cite{attn_uniformer}     & 21.5      & \pzo3.64  & \pzo844   & 121.1         & $224^{2}$ & 82.9  & ICLR'22   \\
      CrossFormer-T~\cite{attn_crossformer} & 27.7      & \pzo2.86  & \pzo948   & 185.4         & $224^{2}$ & 81.5  & ICLR'22   \\
      DAT-T~\cite{attn_DAT}                 & 28.3      & \pzo4.58  & \pzo581   & \pzo75.6      & $224^{2}$ & 82.0  & CVPR'22   \\
      PoolFormer-S36~\cite{attn_metaformer} & 30.8      & \pzo5.00  & \pzo656   & \pzo76.4      & $224^{2}$ & 81.4  & CVPR'22   \\
      MPViT-S~\cite{attn_mpvit}             & 22.8      & \pzo4.80  & \pzo417   & \pzo68.7      & $224^{2}$ & 83.0  & CVPR'22   \\
      Shunted-S~\cite{attn_SSA}             & 22.4      & \pzo5.01  & \pzo461   & \pzo48.6      & $224^{2}$ & 83.7  & CVPR'22   \\
      PoolFormer-S24~\cite{attn_metaformer} & 21.3      & \pzo3.41  & \pzo971   & 111.9         & $224^{2}$ & 80.3  & CVPR'22   \\
      CSWin-T~\cite{cswin}           & 22.3      & \pzo4.34  & \pzo592   & \pzo66.6         & $224^{2}$ & 82.7  & CVPR'22   \\
      iFormer-S~\cite{iformer}           & 19.9      & \pzo4.85  & \pzo528   & \pzo58.8         & $224^{2}$ & 83.4  & NeurIPS'22   \\
      MaxViT-T~\cite{maxvit}           & 25.1      & \pzo4.56  & \pzo367   & \pzo36.6         & $224^{2}$ & 83.6  & ECCV'22   \\
      VAN-Base~\cite{attn_van}              & 26.5      & \pzo5.00  & \pzo542   & \pzo51.9      & $224^{2}$ & 82.8  & arXiv'22  \\
      ViTAEv2-S~\cite{attn_vitae2}         & 19.3      & \pzo5.78  & \pzo464   & \pzo72.8      & $224^{2}$ & 82.6  & IJCV'23  \\
      NAT-Tiny~\cite{attn_nat}              & 27.9      & \pzo4.11  & \pzo401   & < 1           & $224^{2}$ & 83.2  & arXiv'22  \\
      \rowcolor{lblue_tab}
      \textbf{EATFormer-Small}              & 24.3      & \pzo4.32  & \pzo615   & \pzo73.3      & $224^{2}$ & 83.1  & -\pzo\pzo\pzo \\
      \rowcolor{lblue_tab}
      \textbf{EATFormer-Small-384}          & 24.3      & 12.92     & \pzo198   & \pzo24.1      & $384^{2}$ & 84.3  & -\pzo\pzo\pzo \\
      \hline
      \rowcolor{lgray_tab}
      ResNet-101~\cite{resnet,timm_resnet}  & 44.5      & \pzo7.83  & \pzo675   & \pzo81.5      & $224^{2}$ & 81.5  & CVPR'16   \\
      \rowcolor{lgray_tab}
      EfficientNet-B5~\cite{efficientnet}   & 30.3      & 10.46     & \pzo163   & \pzo18.6      & $456^{2}$ & 83.6  & ICML'19   \\
      \rowcolor{lgray_tab}
      PVTv2-B3~\cite{attn_pvt}              & 45.2      & \pzo6.92  & \pzo392   & \pzo45.7      & $224^{2}$ & 83.2  & ICCV'21   \\
      CoaT-Lite Medium~\cite{attn_coat}     & 44.5      & \pzo9.80  & \pzo275   & \pzo42.2      & $224^{2}$ & 83.6  & ICCV'21   \\
      XCiT-S24~\cite{attn_xcit}             & 47.6      & 36.04     & \pzo\pzo99& \pzo16.0      & $224^{2}$ & 82.6  & NeurIPS'21\\
      CSWin-S~\cite{cswin}           & 34.6      & \pzo6.83  & \pzo360   & \pzo51.2         & $224^{2}$ & 83.6  & CVPR'22   \\
      \rowcolor{lblue_tab}
      \textbf{EATFormer-Medium}             & 39.9      & \pzo7.05  & \pzo425   & \pzo53.4      & $224^{2}$ & 83.6  & -\pzo\pzo\pzo \\
      \hline
      \rowcolor{lgray_tab}
      ViT-B/16~\cite{attn_vit}              & 86.5      & 17.58     & \pzo293   & \pzo53.0      & $384^{2}$ & 77.9  & ICLR'21   \\
      \rowcolor{lgray_tab}
      DeiT-B~\cite{attn_deit}               & 86.5      & 17.58     & \pzo293   & \pzo53.7      & $224^{2}$ & 81.8  & ICML'21   \\
      \rowcolor{lgray_tab}
      \textbf{EAT-B}~\cite{eat}             & 86.6      & 14.83     & \pzo331   & \pzo71.5      & $224^{2}$ & 82.0  & NeurIPS'21\\
      \rowcolor{lgray_tab}
      ResNet-152~\cite{resnet,timm_resnet}  & 60.1      & 11.55     & \pzo470   & \pzo57.5      & $224^{2}$ & 82.0  & CVPR'16   \\
      \rowcolor{lgray_tab}
      EfficientNet-B7~\cite{efficientnet}   & 66.3      & 38.32     & \pzo\pzo47& \pzo\pzo5.6   & $600^{2}$ & 84.3  & ICML'19   \\
      \rowcolor{lgray_tab}
      PVTv2-B5~\cite{attn_pvt}              & 81.9      & 11.76     & \pzo256   & \pzo33.9      & $224^{2}$ & 83.8  & ICCV'21   \\
      Swin-B~\cite{attn_swin}               & 87.7      & 15.46     & \pzo258   & \pzo38.6      & $224^{2}$ & 83.5  & ICCV'21   \\
      Twins-SVT-L~\cite{attn_twins}         & 99.2      & 15.14     & \pzo271   & \pzo43.1      & $224^{2}$ & 83.2  & NeurIPS'21\\
      UniFormer-B~\cite{attn_uniformer}     & 49.7      & \pzo8.27  & \pzo378   & \pzo58.7      & $224^{2}$ & 83.9  & ICLR'22   \\
      DAT-B~\cite{attn_DAT}                 & 87.8      & 15.78     & \pzo217   & \pzo30.8      & $224^{2}$ & 84.0  & CVPR'22   \\
      Shunted-B~\cite{attn_SSA}             & 39.6      & \pzo8.18  & \pzo290   & \pzo33.7      & $224^{2}$ & 84.0  & CVPR'22   \\
      PoolFormer-M48~\cite{attn_metaformer} & 73.4      & 11.59     & \pzo301   & \pzo37.5      & $224^{2}$ & 82.5  & CVPR'22   \\
      MPViT-B~\cite{attn_mpvit}             & 74.8      & 16.44     & \pzo181   & \pzo29.1      & $224^{2}$ & 84.0  & CVPR'22   \\
      \blue{CSWin-B}~\cite{cswin}           & \blue{77.4}      & \blue{15.00}  & \blue{\pzo204}   & \blue{\pzo32.5}         & \blue{$224^{2}$} & \blue{84.2}  & \blue{CVPR'22}   \\
      \blue{iFormer-B}~\cite{iformer}           & \blue{47.9}      & \blue{\pzo9.38}  & \blue{\pzo262}   & \blue{\pzo38.9}         & \blue{$224^{2}$} & \blue{84.6}  & \blue{NeurIPS'22}   \\
      \blue{MaxViT-S}~\cite{maxvit}           & \blue{55.8}      & \blue{\pzo9.41}  & \blue{\pzo231}   & \blue{\pzo32.1}         & \blue{$224^{2}$} & \blue{84.4}  & \blue{ECCV'22}   \\
      NAT-Small~\cite{attn_nat}             & 50.7      & \pzo7.50  & \pzo260   & < 1           & $224^{2}$ & 83.7  & arXiv'22  \\
      ViTAEv2-48M~\cite{attn_vitae2}       & 48.6      & 13.38     & \pzo251   & \pzo38.6      & $224^{2}$ & 83.8  & \blue{IJCV'23}  \\		
      \rowcolor{lblue_tab}
      \textbf{EATFormer-Base}               & 49.0      & \pzo8.94  & \pzo329   & \pzo43.7      & $224^{2}$ & 83.9  & -\pzo\pzo\pzo \\
      \rowcolor{lblue_tab}
      \textbf{EATFormer-Base-384}           & 49.0      & 26.11     & \pzo112   & \pzo14.2      & $384^{2}$ & 84.9  & -\pzo\pzo\pzo \\
      \hline
      \end{tabular}
    }
\end{table*}

\subsubsection{Experimental Results}
In this work, we design EATFormer variations at different scales to meet different application requirements, and comparison results with SOTA methods are shown in Table~\ref{table:imagenet}. To fully evaluate the effects of different methods, we choose the number of parameters (Params.), FLOPs, Top-1 accuracy on ImageNet-1K, as well as throughput of GPU (with basic batch size equaling 128 by a single V100 SXM2 32GB, and the batch size will be reduced to the maximum that memory requires for large models) and CPU (with batch size equaling 128 by Xeon 8255C CPU @ 2.50GHz) as evaluation indexes. Our smallest EATFormer-Mobile obtains 69.4 that is much higher than MobileNetV3-Small 0.75$\times$ counterpart, \ie, 65.4, while the largest EATFormer-Base obtains a very competitive result with only 49.0M parameters, and it further achieves 84.9 at 384$\times$384 resolution. \blue{Comparatively, although our approach obtains a slight improvement over recent SOTA MPViT-T/-XS/-S by +0.2\%/+0.0\%/+0.1\%, EATFormer features significantly fewer FLOPs by -0.21G/-0.68G/-0.48G, faster GPU speed by +2.1$\times$/+1.7$\times$/+1.5$\times$, and CPU speed by +1.33$\times$/+1.27$\times$/+1.07$\times$. 
\bblue{At the highest 50M-level model, our EATFormer-B still achieves a throughput of 329 that is 1.8$\times\uparrow$ faster than MPViT-B, and this efficiency increase is also considerable.} 
Meaning that EATFormer is more user-friendly than MPViT on general-purpose GPU and CPU devices, and our EATFormer can better trade-off parameters, computation, and precision.} At the same time, our tiny, small, and base models improve by +5.7$\uparrow$, +2.7$\uparrow$, and +1.9$\uparrow$ compared with the previous conference version. Interestingly, we find that the Top-1 accuracy of different methods with 50$\sim$80M parameters would be approximately saturated to 84.0 without external data, token labeling, larger resolution, \etc, so it is worth future exploration to alleviate this problem.

\begin{table*}[!t]
    \centering
    \caption{Object detection and instance segmentation with Mask R-CNN on COCO~\cite{coco} dataset for 1$\times$ and 3$\times$ schedules. All backbones are pre-trained on ImageNet-1K~\cite{imagenet}.}
    \label{table:det}
    \renewcommand{\arraystretch}{1.0}
    \setlength\tabcolsep{4.0pt}
    \begin{tabular}{p{2.7cm}<{\raggedright} | p{0.56cm}<{\centering} p{0.56cm}<{\centering} p{0.56cm}<{\centering} | p{0.56cm}<{\centering} p{0.56cm}<{\centering} p{0.56cm}<{\centering} | p{0.56cm}<{\centering} p{0.56cm}<{\centering} p{0.56cm}<{\centering} | p{0.56cm}<{\centering} p{0.56cm}<{\centering} p{0.56cm}<{\centering} | p{0.9cm}<{\centering} p{0.8cm}<{\centering} p{1.7cm}<{\raggedleft}}
    \toprule
    \multirow{2}{*}{Backbone} & \multicolumn{6}{c|}{\makecell[c]{Mask R-CNN 1$\times$}} & \multicolumn{6}{c}{\makecell[c]{Mask R-CNN 3$\times$}} & \multirow{2}{*}{\makecell[c]{Params. \\(M)$\downarrow$}} & \multirow{2}{*}{\makecell[c]{FLOPs \\(G)$\downarrow$}} & \multirow{2}{*}{\makecell[c]{Pub.\pzo\pzo}} \\
    \cmidrule(lr){2-7} \cmidrule(lr){8-13}
    & $AP^b$ & $AP^b_{50}$ & $AP^b_{75}$ & $AP^m$ & $AP^m_{50}$ & $AP^m_{75}$ & $AP^b$ & $AP^b_{50}$ & $AP^b_{75}$ & $AP^m$ & $AP^m_{50}$ & $AP^m_{75}$ & & & \\
    \midrule
    PVT-Tiny~\cite{attn_pvt} & 36.7 & 59.2 & 39.3 & 35.1 & 56.7 & 37.3 & 39.8 & 62.2 & 43.0 & 37.4 & 59.3 & 39.9 & 33 & - & ICCV'21 \\
    PVTv2-B0~\cite{attn_pvt2} & 38.2 & 60.5 & 40.7 & 36.2 & 57.8 & 38.6 & - & - & - & - & - & - & 23 & 195 & CVM'22 \\
    XCiT-T12~\cite{attn_xcit} & - & - & - & - & - & - & 44.5 & 66.4 & 48.8 & 40.4 & 63.5 & 43.3 & 26 & 266 & NeurIPS'21 \\
    PFormer-S12~\cite{attn_metaformer} & 37.3 & 59.0 & 40.1 & 34.6 & 55.8 & 36.9 & - & - & - & - & - & - & 31 & - & CVPR'22 \\
    MPViT-T~\cite{attn_mpvit} & 42.2 & 64.2 & 45.8 & 39.0 & 61.4 & 41.8 & 44.8 & 66.9 & 49.2 & 41.0 & 64.2 & 44.1 & 28 & 216 & CVPR'22 \\
    \rowcolor{lblue_tab}
    EATFormer-Tiny & 42.3 & 64.7 & 46.2 & 39.0 & 61.5 & 42.0 & 45.4 & 67.5 & 49.5 & 41.4 & 64.8 & 44.6 & 25 & 198 & -\pzo\pzo\pzo \\
    \hline
    ResNet-50~\cite{resnet} & 38.0 & 58.6 & 41.4 & 34.4 & 55.1 & 36.7 & 41.0 & 61.7 & 44.9 & 37.1 & 58.4 & 40.1 & 44 & 260 & CVPR'16 \\
    Swin-T~\cite{attn_swin} & 43.7 & 66.6 & 47.7 & 39.8 & 63.3 & 42.7 & 46.0 & 68.1 & 50.3 & 41.6 & 65.1 & 44.9 & 48 & 267 & ICCV'21 \\
    Twins-S~\cite{attn_twins} & 43.4 & 66.0 & 47.3 & 40.3 & 63.2 & 43.4 & 46.8 & 69.2 & 51.2 & 42.6 & 66.3 & 45.8 & 44 & 228 & NeurIPS'21 \\
    PFormer-S24~\cite{attn_metaformer} & 40.1 & 62.2 & 43.4 & 37.0 & 59.1 & 39.6 & - & - & - & - & - & - & 41 & - & CVPR'22 \\
    DAT-T~\cite{attn_DAT} & 44.4 & 67.6 & 48.5 & 40.4 & 64.2 & 43.1 & 47.1 & 69.2 & 51.6 & 42.4 & 66.1 & 45.5 & 48 & 272 & CVPR'22 \\  
    \blue{MPViT-S}~\cite{attn_mpvit} & \blue{46.4} & \blue{68.6} & \blue{51.2} & \blue{42.4} & \blue{65.6} & \blue{45.7} & \blue{48.4} & \blue{70.5} & \blue{52.6} & \blue{43.9} & \blue{67.6} & \blue{47.5} & \blue{43} & \blue{268} & \blue{CVPR'22} \\
    \rowcolor{lblue_tab}
    EATFormer-Small & 46.1 & 68.4 & 50.4 & 41.9 & 65.3 & 44.8 & 47.4 & 69.3 & 51.9 & 42.9 & 66.4 & 46.3 & 44 & 258 & -\pzo\pzo\pzo \\
    \hline
    ResNet-101~\cite{resnet} & 40.4 & 61.1 & 44.2 & 36.4 & 57.7 & 38.8 & 42.8 & 63.2 & 47.1 & 38.5 & 60.1 & 41.3 & 63 & 336 & CVPR'16 \\
    Swin-S~\cite{attn_swin} & 45.7 & 67.9 & 50.4 & 41.1 & 64.9 & 44.2 & 48.5 & 70.2 & 53.5 & 43.3 & 67.3 & 46.6 & 69 & 359 & ICCV'21 \\
    Twins-B~\cite{attn_twins} & 45.2 & 67.6 & 49.3 & 41.5 & 64.5 & 44.8 & 48.0 & 69.5 & 52.7 & 43.0 & 66.8 & 46.6 & 76 & 340 & NeurIPS'21 \\
    PFormer-S36~\cite{attn_metaformer} & 41.0 & 63.1 & 44.8 & 37.7 & 60.1 & 40.0 & - & - & - & - & - & - & 51 & - & CVPR'22 \\
    DAT-S~\cite{attn_DAT} & 47.1 & 69.9 & 51.5 & 42.5 & 66.7 & 45.4 & 49.0 & 70.9 & 53.8 & 44.0 & 68.0 & 47.5 & 69 & 378 & CVPR'22 \\
    \blue{MPViT-B}~\cite{attn_mpvit} & \blue{48.2} & \blue{70.0} & \blue{52.9} & \blue{43.5} & \blue{67.1} & \blue{46.8} & \blue{49.5} & \blue{70.9} & \blue{54.0} & \blue{44.5} & \blue{68.3} & \blue{48.3} & \blue{95} & \blue{503} & \blue{CVPR'22} \\
    \rowcolor{lblue_tab}
    EATFormer-Base & 47.2 & 69.4 & 52.1 & 42.8 & 66.4 & 46.5 & 49.0 & 70.3 & 53.6 & 44.2 & 67.7 & 47.6 & 68 & 349 & -\pzo\pzo\pzo \\
    \hline
    \end{tabular}
\end{table*}

\subsection{Object Detection and Instance Segmentation}
\subsubsection{Experimental Setting}
To further evaluate the effectiveness and superiority of our method, ImageNet-1K~\cite{imagenet} pre-trained EATFormer is benchmarked as the feature extractor for downstream object detection and instance segmentation tasks on COCO2017 dataset~\cite{coco}, and its window size increases from 7 to 12 without global attention and other changes. For fair comparisons, we employ MMDetection library~\cite{mmdetection} for experiments and follow the same training recipe as Swin-Transformer~\cite{attn_swin}: 1$\times$ schedule for 12 epochs and 3$\times$ schedule with a multi-scale training strategy for 36 epochs. AdamW~\cite{adamw} optimizer is used for training with learning rate and weight decay equaling 1$e^{-4}$ and 5$e^{-2}$, respectively. 

\begin{figure}[!t]
    \centering
    \includegraphics[width=1.0\linewidth]{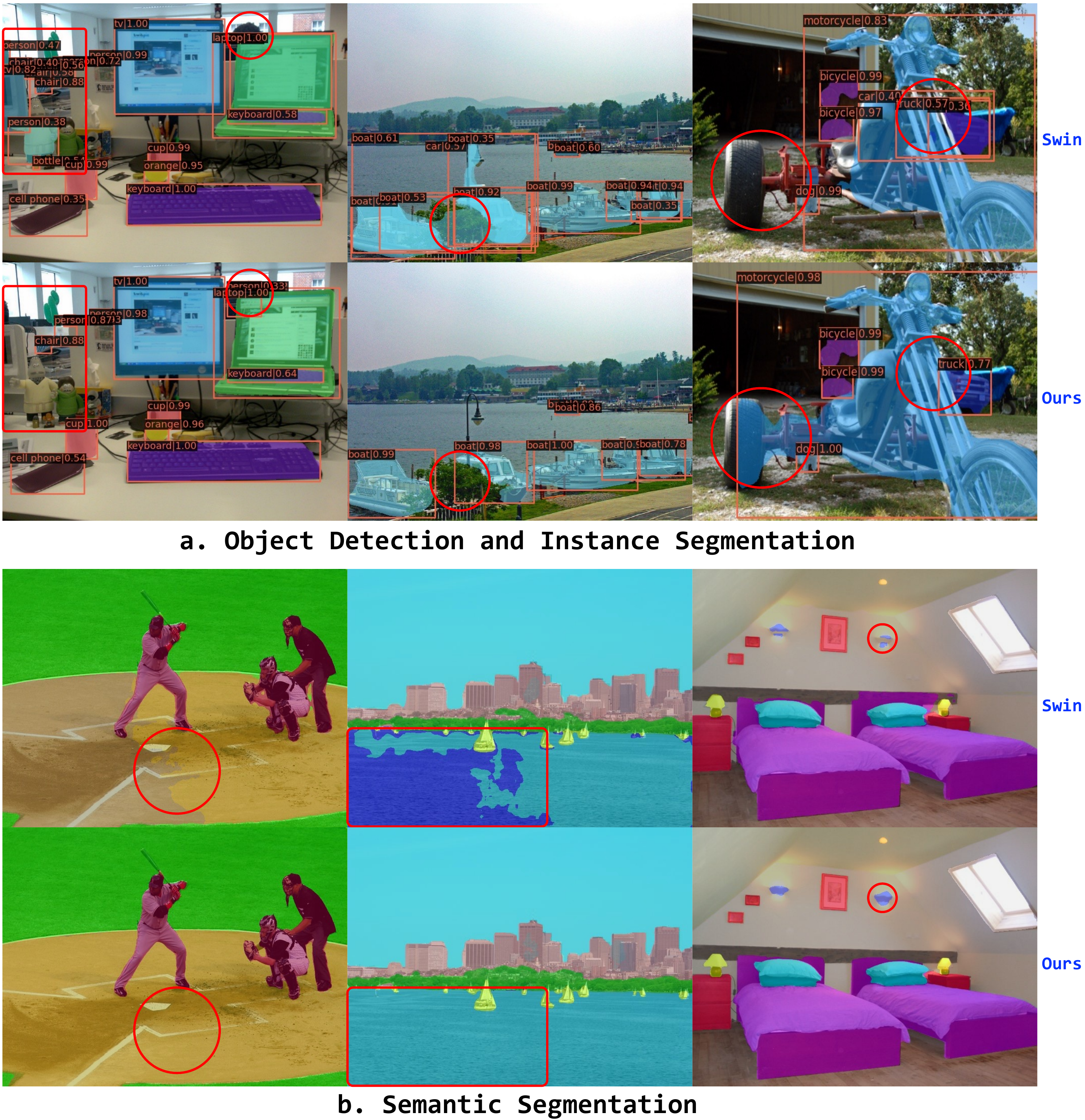}
    \caption{Intuitive visualizations of two downstream tasks compared with Swin Transformer. Distinct differences are highlighted with red circles and rectangles.} 
    \label{fig:explanation_downstream}
\end{figure}

\subsubsection{Experimental Results}
Comparison results of box mAP ($AP^b$) and mask mAP ($AP^m$) are reported in Table~\ref{table:det}, and our improved EATFormer obtains competitive results over recent approaches. Specifically, our tiny model obtains +5.6$\uparrow$/+5.6$\uparrow$ $AP^b$ and +3.9$\uparrow$/+4.0$\uparrow$ $AP^m$ improvements over PVT-Tiny~\cite{attn_pvt} on both 1$\times$ and 3$\times$ schedules, while achieves higher results over MPViT~\cite{attn_mpvit} with less parameters and FLOPs, \ie, +0.6$\uparrow$ and +0.4$\uparrow$ on 3$\times$ schedule. For larger EATFormer-small and EATFormer-base models, we consistently get better results than recent counterparts, which surpass Swin-T by +2.4$\uparrow$/+2.1$\uparrow$ and Swin-S by +1.5$\uparrow$/+1.7$\uparrow$ with 1$\times$ schedule, while by +1.4$\uparrow$/+1.3$\uparrow$ and by +0.5$\uparrow$/+0.9$\uparrow$ with 3$\times$ schedule. Also, we obtain slightly higher results than DAT~\cite{attn_DAT} with computation amount going down by 29G$\downarrow$. Although EATFormer is slightly lower than SOTA MPViT-S/B in the downstream task metrics, our method has obvious advantages in the number of parameters and computation, \eg, -10G$\downarrow$ FLOPs decreasing than MPViT-S for Mask R-CNN, while -154G$\downarrow$ (-30.6\%) FLOPs and -27M$\downarrow$ (-28.4\%) parameters decreasing than MPViT-B, \bblue{effectively balancing the trade-off between effectiveness and performance.} For MPViT-T, our EATFormer-Tiny has obvious metrics, parameter numbers, and computation advantages. Qualitative visualizations on validation dataset compared with Swin-S~\cite{attn_swin} are shown in the top part of Figure~\ref{fig:explanation_downstream}. Results indicate that our EATFormer can obtain more accurate detection accuracy, fewer false positives, and finer segmentation results than Swin Transformer.

\subsection{Semantic Segmentation}
\subsubsection{Experimental Setting}
We further conduct semantic segmentation experiments on the ADE20K~\cite{ade20k} dataset, and pre-trained EATFormer with window size equaling 12 is integrated into UperNet~\cite{upernet} architecture to obtain pixel-level predictions. In detail, we follow the same setting of Swin-Transformer~\cite{attn_swin} to train the model for 160k iterations. AdamW~\cite{adamw} optimizer is also used with learning rate and weight decay equaling 1$e^{-4}$ and 5$e^{-2}$, respectively. 

\subsubsection{Experimental Results}
Segmentation results compared with contemporary SOTA works under three main model scales are reported in Table~\ref{table:seg}. Our EATFormer-Tiny obtains a significantly +3.4$\uparrow$ improvement than recent VAN-Tiny~\cite{attn_van}, while EATFormer-Small achieves a higher mIoU with fewer FLOPs over SOTA methods. For larger EATFormer-Base, it consistently obtains competitive results, \ie, +1.7$\uparrow$ and +1.0$\uparrow$ than Swin-S~\cite{attn_swin} and DAT-S~\cite{attn_DAT}, respectively. Compared with SOTA MPViT, we obtain a better trade-off among parameters, computation, and precision. \bblue{\Eg, our EATFormer-Base has 26M fewer parameters and 156G fewer FLOPs compared to MPViT-B.} 
Our approach generally has excellent overall precision and computation performance than counterpart. Also, intuitive visualizations of the validation dataset compared with Swin-S~\cite{attn_swin} are shown in the bottom part of Figure~\ref{fig:explanation_downstream}. Qualitative results consistently demonstrate the robustness and effectiveness of the proposed approach, where our EATFormer has more accurate segmentation results.

\begin{table}[!t]
    \centering
    \caption{Semantic segmentation results compared with SOTAs on ADE20K~\cite{ade20k} by Upernet~\cite{upernet}.}
    \label{table:seg}
    \renewcommand{\arraystretch}{1.0}
    \setlength\tabcolsep{3.0pt}
    \begin{tabular}{p{2.6cm}<{\raggedright} p{0.9cm}<{\centering} p{1.0cm}<{\centering} p{1.0cm}<{\centering} p{1.7cm}<{\raggedleft}}
    \toprule
    Backbone & Params. & GFLOPs & mIoU & Pub.\pzo\pzo \\
    \hline
    XCiT-T12~\cite{attn_xcit}           & \pzo34    & \pzo-     & 43.5  & NeurIPS'21    \\
    VAN-Tiny~\cite{attn_van}            & \pzo32    & \pzo858   & 41.1  & CVPR'22       \\
    \rowcolor{lblue_tab}
    EATFormer-Tiny                      & \pzo34    & \pzo870   & 44.5  & -\pzo\pzo\pzo \\
    \hline
    Swin-T~\cite{attn_swin}             & \pzo60    & \pzo945   & 44.5  & ICCV'21       \\
    XCiT-S12~\cite{attn_xcit}           & \pzo52    & \pzo-     & 46.6  & NeurIPS'21    \\
    DAT-T~\cite{attn_DAT}               & \pzo60    & \pzo957   & 45.5  & CVPR'22       \\
    ViTAEv2-S~\cite{attn_vitae2}        & \pzo49    & \pzo-     & 45.0  & \blue{IJCV'23}      \\
    \blue{MPViT-S}~\cite{attn_mpvit}           & \blue{\pzo52}    & \blue{\pzo943}   & \blue{48.3}  & \blue{CVPR'22}      \\
    UniFormer-S~\cite{uniformer_arxiv}  & \pzo52    & \pzo955   & 47.0  & arXiv'22      \\
    \rowcolor{lblue_tab}
    EATFormer-Small                     & \pzo53    & \pzo934   & 47.3  & -\pzo\pzo\pzo \\
    \hline
    Swin-S~\cite{attn_swin}             & \pzo81    & 1038      & 47.6  & ICCV'21       \\
    XCiT-M24~\cite{attn_xcit}           & 109       & \pzo-     & 48.4  & NeurIPS'21    \\
    DAT-S~\cite{attn_DAT}               & \pzo81    & 1079      & 48.3  & CVPR'22       \\
    \blue{MPViT-B}~\cite{attn_mpvit}           & \blue{105}       & \blue{1186}      & \blue{50.3}  & \blue{CVPR'22}      \\
    UniFormer-B~\cite{uniformer_arxiv}  & \pzo80    & 1106      & 49.5  & arXiv'22      \\
    \rowcolor{lblue_tab}
    EATFormer-Base                      & \pzo79    & 1030      & 49.3  & -\pzo\pzo\pzo \\
    \hline
    \end{tabular}
\end{table}

\subsection{Ablation Study}
To fully evaluate the effectiveness of each designed module, we conduct a series of ablation studies in the following sections. By default, EATFormer-Tiny is used for all experiments, and we follow the same training recipe as mentioned in Section~\ref{section:exp_cls}.

\subsubsection{Component of EAT Block} \label{ablation:eat}
As afore-mentioned in Section~\ref{section:eatformer}, our proposed EAT block contains: \emph{1)} MSRA, \emph{2)} GLI, and \emph{3)} FFN modules that are responsible for aggregating multi-scale information, interacting global and local features, and enhancing the features of each location, respectively. To verify the validity of each module in the EAT block, we conduct an ablation experiment in Table~\ref{tab:ablation_block} that contains different component combinations. Results indicate that each component contributes to the model performance, and our EATFormer obtains the best result when using all three parts. Since FFN takes up most of the parameters and calculations, we can conduct further research on optimizing this module to obtain better-integrated model performance.
 
\begin{table}[t]
    \centering
    \renewcommand\arraystretch{0.9}
    \setlength\tabcolsep{5pt}
    \begin{tabular}{C{25pt}C{25pt}C{25pt}C{28pt}C{28pt}C{28pt}}
       \toprule
       MSRA & GLI & FFN & Params & FLOPs & Top-1\\
       \midrule
       \cmark   & \xmarkg   & \xmarkg   & 2.4   & 0.45  & 62.9  \\
       \xmarkg  & \cmark    & \xmarkg   & 2.6   & 0.51  & 64.4  \\
       \xmarkg  & \xmarkg   & \cmark    & 5.2   & 1.17  & 71.4  \\
       \cmark   & \cmark    & \xmarkg   & 2.9   & 0.60  & 67.7  \\
       \cmark   & \xmarkg   & \cmark    & 5.5   & 1.26  & 76.0  \\
       \xmarkg  & \cmark    & \cmark    & 5.8   & 1.32  & 77.4  \\
       \cmark   & \cmark    & \cmark    & 6.1   & 1.41  & 78.4  \\
       \bottomrule
    \end{tabular}
    \caption{Ablation study for different component combinations in EAT block.}
    \label{tab:ablation_block}
\end{table}

\subsubsection{Separation Ratio of GLI} \label{ablation:ratio}
\begin{figure}[!t]
    \centering
    \includegraphics[width=1.0\linewidth]{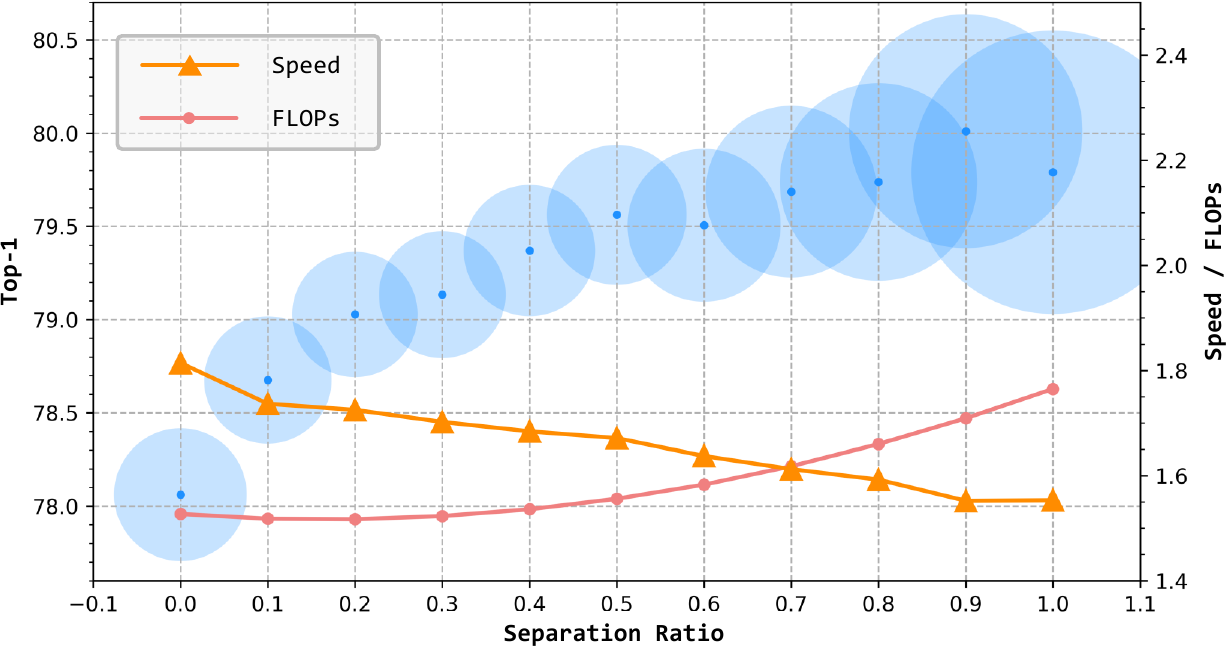}
    \caption{Separation ratio analysis of GLI. \textcolor[RGB]{30,145,255}{\textbf{Blue}} circle represents the Top-1 accuracy of EATFormer at different separation ratios (\cf, left axis), and the radius represents the relative number of parameters. \textcolor[RGB]{255,140,0}{\textbf{Orange}} and \textcolor[RGB]{240,128,128}{\textbf{pink}} lines represent the running speed and relative FLOPs of the model (\cf, right axis).}
    \label{fig:ratio}
\end{figure}

We deduce from Equation~\ref{eq:params} and Equation~\ref{eq:flops} in Section~\ref{section:gli} that EATFormer has the lowest number of parameters and calculation amount when separation ratio $p$ of GLI equals 0.2, and there is not much difference about the total parameters and calculations when $p$ lies in the range [0, 0.5]. To further prove the above analysis and verify the validity of the GLI, we conduct a set of experiments with equal interval sampling of $p$ in range [0, 1] for the classification task. As shown in Figure~\ref{fig:ratio}, the x-coordinate represents different proportions, and the left y-ordinate represents Top-1 accuracy of the modified EATFormer-Tiny with embedding dims equaling [64, 128, 230, 320] for divisible channels. The right y-ordinate shows the model's running speed and relative computation amount. Results in the figure are consistent with the foregoing derivation, and $p$ equaling 0.5 is the most economical and efficient choice, where the model has relatively high precision, fast speed, and low computational cost. All GLI layers in this article use the same ratio, and exploring different ratios for different layers should lead to further improvements based on the above analysis. 

\subsubsection{Component Ablation of EATFormer} \label{ablation:eatformer}
Following the core idea of paralleling global and local modelings, this paper extends a pyramid architecture over the previous columnar EAT model~\cite{eat}. Specifically, EAT block-based EATFormer can be seen as evolving from the naive baseline, which employs: \emph{1)} patch embedding for down-sampling; \emph{2)} MSRA with only one scale; \emph{3)} naive MSA ; \emph{4)} simple
addition operation with $\alpha_{i}, i=1, \dots, N, g, l$ equaling 1, instead of: \emph{1)} MSRA for down-sampling ; \emph{2)} MSRA with multiple scale; \emph{3)} improved MD-MSA ; \emph{4)} weighted operation mixing (WOM) with learnable $\alpha_{i}, i=1, \dots, N, g, l$. Detail ablation experiment based on EATFormer-tiny can be viewed in Table~\ref{tab:ablation_eaformer}, and the results indicate that each individual component has a role, and different components combination can complement each other to help the model achieve higher results. Note that WOM can only be applied if multi-path-based MSRA is used.

\begin{table}[t]
    \centering
    \renewcommand\arraystretch{0.9}
    \setlength\tabcolsep{3pt}
    \begin{tabular}{C{25pt}C{25pt}C{25pt}C{28pt}C{28pt}C{28pt}C{28pt}}
       \toprule
       \multicolumn{1}{c}{\multirow{2}{*}{\makecell[c]{MSRA\\ Down}}} & \multicolumn{1}{c}{\multirow{2}{*}{MSRA}} & \multicolumn{1}{c}{\multirow{2}{*}{\makecell[c]{MD-\\MSA}}} & \multicolumn{1}{c}{\multirow{2}{*}{WOM}} & \multicolumn{1}{c}{\multirow{2}{*}{Param.}} & \multicolumn{1}{c}{\multirow{2}{*}{FLOPs}} & \multicolumn{1}{c}{\multirow{2}{*}{Top-1}} \\
       &  &  &  &  &  & \\
       \midrule
       \xmarkg  & \xmarkg   & \xmarkg   & \xmarkg   & 4.792   & 1.232  & 77.4 \\
       \cmark   & \xmarkg   & \xmarkg   & \xmarkg   & 5.202   & 1.300  & 77.8 \\
       \xmarkg  & \cmark    & \xmarkg   & \xmarkg   & 5.208   & 1.283  & 77.9 \\
       \xmarkg  & \xmarkg   & \pmark    & \xmarkg   & 4.804   & 1.236  & 77.5 \\
       \xmarkg  & \xmarkg   & \cmark    & \xmarkg   & 4.805   & 1.236  & 77.7 \\
       \cmark   & \cmark    & \xmarkg   & \xmarkg   & 6.109   & 1.412  & 78.2 \\
       \cmark   & \xmarkg   & \cmark    & \xmarkg   & 5.214   & 1.304  & 78.0 \\
       \xmarkg  & \cmark    & \cmark    & \xmarkg   & 5.220   & 1.288  & 78.1 \\
       \cmark   & \cmark    & \cmark    & \xmarkg   & 6.122   & 1.416  & 78.2 \\
       \cmark   & \cmark    & \xmarkg   & \cmark    & 6.109   & 1.412  & 78.1 \\
       \xmarkg  & \cmark    & \cmark    & \cmark    & 5.221   & 1.288  & 78.0 \\
       \cmark   & \cmark    & \cmark    & \cmark    & 6.122   & 1.416  & 78.4 \\
       \bottomrule
    \end{tabular}
    \caption{Ablation study for different component combinations in EATFormer. \cmark means choice while \xmarkg is the opposite, and \pmark represents D-MSA that abandons the modulation operation.}
    \label{tab:ablation_eaformer}
\end{table}

\subsubsection{Composition of GLI} \label{ablation:gli}
By default, the global path in GLI employs the designed MD-MSA module inspired by the dynamic population concept, while the local branch uses conventional CNN to model static feature extraction. To further assess the potential of the GLI module, different combinations of global (\ie, MSA and MD-MSA) and local (\ie, CNN and DCNv2~\cite{dcn2}) operators are used for experiments. As shown in Table~\ref{tab:ablation_gli}, MD-MSA improves the model effect by $0.3\uparrow$ only with negligible parameters and computation, while DCNv2 can further boost performance by a large margin at the cost of higher storage and computation. Theoretically, MD-MSA has no significant impact on the speed, but the naive PyTorch implementation without CUDA acceleration leads to a obvious decrease in GPU speed. Therefore, the running speed of our model could be improved after further optimization for MD-MSA.

\begin{table}[t]
    \centering
    \renewcommand\arraystretch{0.9}
    \setlength\tabcolsep{3.6pt}
    \begin{tabular}{C{45pt}C{30pt}C{30pt}C{30pt}C{28pt}C{28pt}}
       \toprule
       Global & Local & Param. & FLOPs & GPU & Top-1 \\
       \midrule
       MSA      & CNN   & \pzo6.1   & 1.412 & 1896      & 78.1  \\
       MSA      & DCNv2 & \pzo9.0   & 1.522 & 1567      & 79.0  \\
       MD-MSA   & CNN   & \pzo6.1   & 1.416 & 1549      & 78.4  \\
       MD-MSA   & DCNv2 & \pzo9.0   & 1.526 & 1333      & 79.2  \\
       \bottomrule
    \end{tabular}
    \caption{Ablation study for compositions of GLI.}
    \label{tab:ablation_gli}
\end{table}

\subsubsection{Normalization Type} \label{ablation:norm}
Transformer-based vision models generally use \emph{Layer Normalization} (LN) to achieve better results rather than \emph{Batch Normalization} (BN). Nevertheless, considering that LN requires slightly more computation than BN and the proposed hybrid EATFormer contains many convolutions that are usually combined with \emph{Batch Normalization} (BN) layers, we conduct an ablation study to evaluate which normalization would be better. Table~\ref{tab:ablation_norm} shows the results on three EATFormer variants, and BN-normalized EATFormer achieves slightly better results while owing an significantly faster GPU inference speed. Note that merging convolution and BN layers is not used here, and this technique can further improve the inference speed.

\begin{table}[t]
    \centering
    \renewcommand\arraystretch{0.9}
    \setlength\tabcolsep{5pt}
    \begin{tabular}{C{40pt}C{30pt}C{30pt}C{30pt}C{28pt}}
       \toprule
       Network & Params & FLOPs & GPU & Top-1  \\
       \midrule
       Tiny-LN      & \pzo6.1   & 1.425 & \pzo963   & 78.2  \\
       Tiny-BN      & \pzo6.1   & 1.416 & 1549      & 78.4  \\
       \midrule
       Small-LN     & 24.3      & 4.337 & \pzo448   & 82.8  \\
       Small-BN     & 24.3      & 4.320 & \pzo615   & 83.1  \\
       \midrule
       Base-LN      & 49.0      & 8.775 & \pzo240   & 83.7  \\
       Base-BN      & 49.0      & 8.744 & \pzo345   & 83.9  \\
       \bottomrule
    \end{tabular}
    \caption{Effects of different normalization types.}
    \label{tab:ablation_norm}
\end{table}

\subsubsection{MSRA at Different Stages} \label{ablation:msra}
Different network depths may have different requirements for the MSRA module, so we explore the introduction of MSRA at different stages. As shown in Table~\ref{tab:ablation_msra}, our model obtains the best result when MSRA is used in [2, 3, 4] stages, and the model effect decreases sharply when only used in the fourth stage. Considering the model accuracy and efficiency, using this module in [3, 4] stages is a better choice.

\begin{table}[t]
    \centering
    \renewcommand\arraystretch{0.9}
    \setlength\tabcolsep{5pt}
    \begin{tabular}{C{40pt}C{30pt}C{30pt}C{30pt}C{28pt}}
       \toprule
       Stages & Params & FLOPs & GPU & Top-1  \\
       \midrule
       $[$1, 2, 3, 4$]$     & 6.3   & 1.541 & 1291  & 78.2  \\
       $[$2, 3, 4$]$        & 6.3   & 1.533 & 1434  & 78.5  \\
       $[$3, 4$]$           & 6.1   & 1.416 & 1549  & 78.4  \\
       $[$4$]$              & 5.6   & 1.326 & 1695  & 77.9  \\
       \bottomrule
    \end{tabular}
    \caption{Ablation study of MSRA on different stages.}
    \label{tab:ablation_msra}
\end{table}

\subsubsection{Kernel Size of MSRA} \label{ablation:msra_kernel}
The MSRA module for multi-scale modeling adopts CNN as its primary component so that the convolution kernel may influence the model results. As shown in Table~\ref{tab:ablation_kernel}, a larger kernel size can only slightly increase the model effect, but the number of parameters and the amount of calculation could increase dramatically. Therefore, we employ efficient $3 \times 3$ kernel size in MSRA for EATFormer at all scales. 

\begin{table}[t]
    \centering
    \renewcommand\arraystretch{0.9}
    \setlength\tabcolsep{5pt}
    \begin{tabular}{C{40pt}C{30pt}C{30pt}C{30pt}C{28pt}}
       \toprule
       Size & Params & FLOPs & GPU & Top-1  \\
       \midrule
       3$\times$3   & \pzo6.1   & 1.416 & 1549  & 78.4  \\
       5$\times$5   & \pzo9.0   & 1.845 & 1342  & 78.5  \\
       7$\times$7   & 13.4      & 2.487 & 1087  & 78.5  \\
       \bottomrule
    \end{tabular}
    \caption{Ablation study on kernel size of MSRA.}
    \label{tab:ablation_kernel}
\end{table}

\subsubsection{Layer Number of TRH} \label{ablation:trh}
The Plug-and-play TRH module can easily be docked with the transformer backbone to obtain the task-related feature representation, and we take the classification task as an example to explore the effect of this module. As shown in Table~\ref{tab:ablation_trh}, Top-1 accuracy is significantly improved by gradually increasing the number of TRH layers in the EATFormer-Tiny model, and the performance tends to saturation after two layers. Therefore, using two-layer TRH is the recommended choice to balance model effectiveness and efficiency. However, there is no noticeable improvement in the larger models, so the multi-task advantage of TRH for the larger model is more important than accuracy improvement.

\begin{table}[t]
    \centering
    \renewcommand\arraystretch{0.9}
    \setlength\tabcolsep{5pt}
    \begin{tabular}{p{1.7cm}<{\raggedright} p{1.0cm}<{\raggedright} p{1.0cm}<{\centering} p{1.0cm}<{\centering} p{1.0cm}<{\centering}}
      \toprule
      Network     & Params            & FLOPs & GPU     & Top-1 \\
      \midrule
      Tiny        & \pzo6.1           & 1.416 & 1549    & 78.4  \\
      Tiny~+1     & \pzo6.9$_{+0.8}$  & 1.423 & 1495    & 78.7  \\
      Tiny~+2     & \pzo7.7$_{+1.6}$  & 1.430 & 1461    & 79.1  \\
      Tiny~+3     & \pzo8.4$_{+2.3}$  & 1.438 & 1423    & 79.2  \\
      \midrule
      Small~+2    & 29.1$_{+4.8}$     & 4.363 & \pzo589 & 83.2  \\
      Base~+2     & 55.3$_{+6.3}$     & 9.001 & \pzo316 & 83.9  \\
      \bottomrule
    \end{tabular}
    \caption{Quantitative ablation study for the layer number of TRH.}
    \label{tab:ablation_trh}
\end{table}

\subsection{EATFormer Explanation}

\subsubsection{Alpha Distribution of Different Depths} \label{explanation:alpha}
\begin{figure}[!t]
    \centering
    \includegraphics[width=1.0\linewidth]{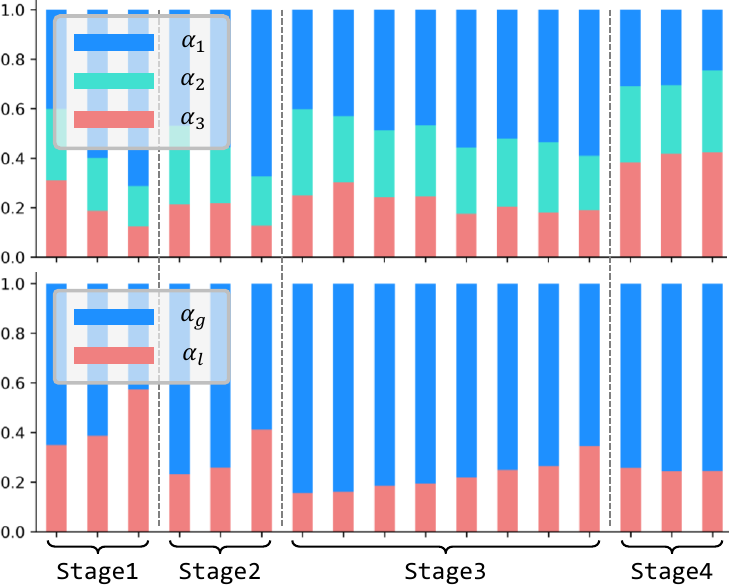}
    \caption{Alpha distribution of the trained EATFormer for different depths. Top and bottom parts represent $\alpha_{i}, i=1, \dots, N$ in MSRA and $\alpha_{i}, i=g, l$ in GLI, respectively.}
    \label{fig:explanation_alpha}
\end{figure}

The weighted operation mixing mechanism can improve the model performance and objectively represent the model's attention to different branches at different depths. Based on EATFormer-tiny, we use 3-path MSRA along with 2-path GLI for each EAT block, and the alpha-indicated weight distribution after training is shown in Figure~\ref{fig:explanation_alpha}. \emph{1)} For the MSRA module, the proportion of $\alpha_{1}$ (\ie, dilation equals 1) in the same stage shows an increasing trend while the larger $\alpha_{3}$ is the opposite, indicating that local feature extraction with stronger correlation (\ie, smaller scale) is more critical for the network. And weight mutation between adjacent stages is caused by a down-sampling operation that changes the feature distribution. In the last stage4, large scale paths have more weight because they need to model as much global information as possible to get proper classification results. But in general, the proportion of each branch is balanced, meaning that feature learning at all scales contributes to the network. Considering the amount of computation and the number of parameters, this also supports the experimental result about why only using MSRA for stage3/4 described in above Section~\ref{ablation:msra}. \emph{2)} For the GLI module, the global branch has more and more weight than the local branches as the network deepens, indicating that both branches are effective and complement each other: local CNN is more suitable for low-level feature extraction while the global transformer is better at high-level information fusion.

\subsubsection{Attention Visualization} \label{explanation:cam}
\begin{figure}[!t]
    \centering
    \includegraphics[width=1.0\linewidth]{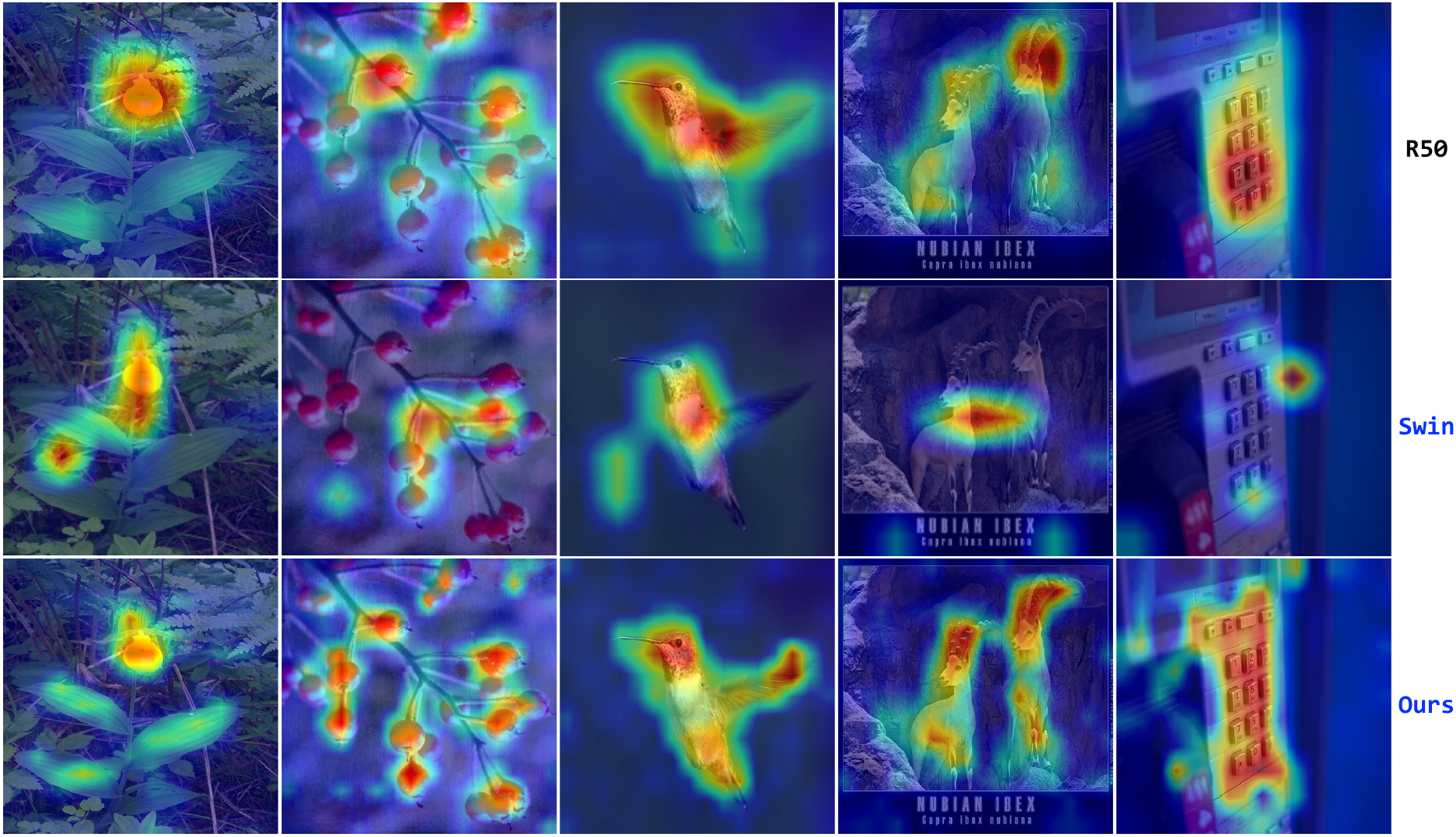}
    \caption{Attention visualizations by Grad-CAM of our EATFormer compared to CNN-based ResNet-50~\cite{resnet} and transformer-based Swin-B~\cite{attn_swin}.}
    \label{fig:explanation_cam}
\end{figure}

To better illustrate which parts of the image the model focuses on, Grad-CAM~\cite{gradcam} is applied to highlight concerning regions by our small model. As shown in Figure~\ref{fig:explanation_cam}, we visualize different images by column for ResNet-50~\cite{resnet}, Swin-B~\cite{attn_swin}, and our EATFormer-Base models, respectively. Results indicate that: \emph{1)} CNN-based ResNet tends to focus on as many regions as possible but ignores edges; \emph{2)} Transformer-based Swin pays more attention to sparse local areas; \emph{3)} Thanks to the design of MSRA and GLI modules, our EATFormer has more discriminative attention to subject targets that own very sharp edges.

\subsubsection{Attention Distance of Global Path in GLI} \label{explanation:gli}
\begin{figure}[!t]
    \centering
    \includegraphics[width=1.0\linewidth]{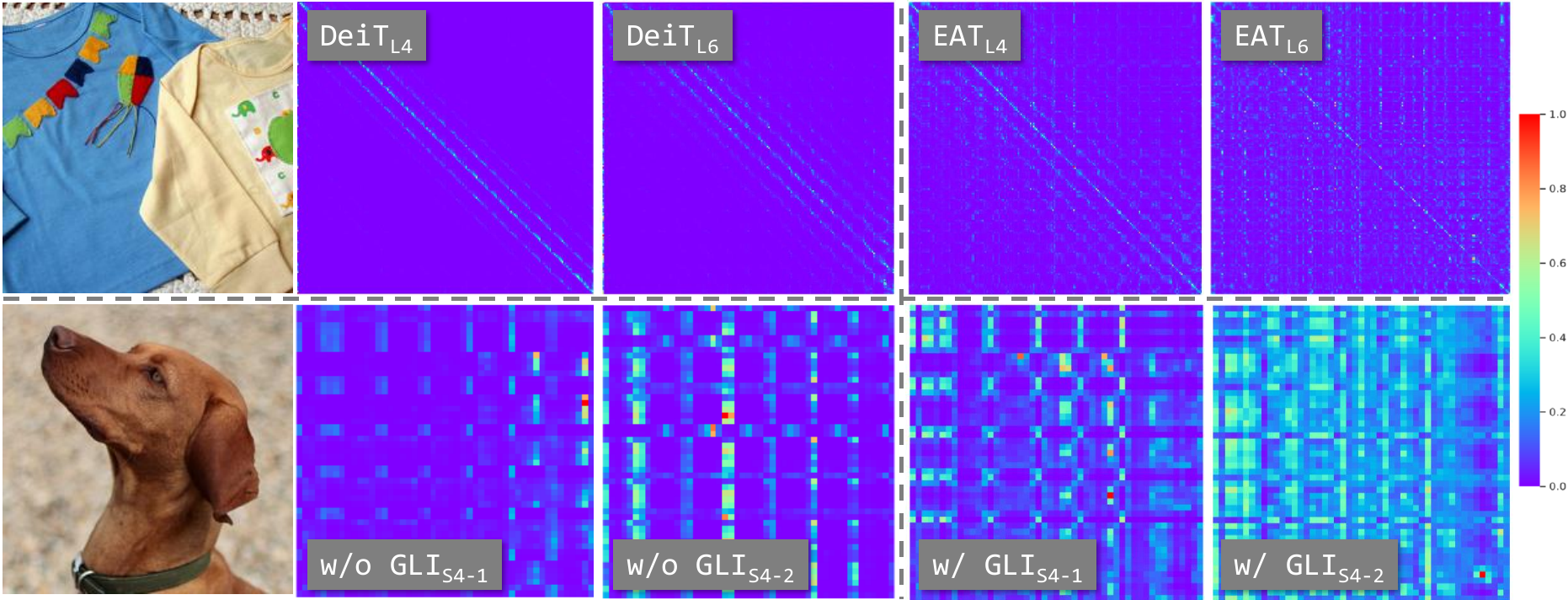}
    \caption{Attention scope for the transformer-based global branch in GLI. The top part shows the attention maps for columnar DeiT~\cite{attn_deit} (w/o local modeling) and our previous EAT model~\cite{eat} (w/ local modeling) in different depths; the bottom part shows results of EATFormer w/ and w/o local modelings in the last stage.}
    \label{fig:explanation_gli}
\end{figure}

We design the GLI module to explicitly model global and local information separately, so the local branch could undertake part of the short-distance modeling of the global branch. To verify this, we visualize the modeling distance of the global branch for our previous columnar EAT model~\cite{eat} and current studied EATFormer in Figure~\ref{fig:explanation_gli}: \emph{1-Top)} Compared with DeiT without local modeling, our EAT pays more attention to global information fusion (choosing layer 4/6 for examples), where more significant values are found at off-diagonal locations. \emph{2-Bottom)} Attention maps in the last stage are visualized because the window size equals the feature size that could cover overall information. When using global modeling alone (w/o GLI), the model only focuses on sparse regions but will pay attention to more regions when GLI is used. Results indicate that the designed parallel local path takes responsibility for some local modelings that should be the responsibility of the global path. We can find differences in feature modeling between columnar-alike and pyramid-aware architectures.

\blue{\textbf{Relationship with EA.} Motivated by EA variants~\cite{mea1,mea2,mea3} that introduce local search procedures besides conventional global search for converging higher-quality solutions, we analogically improve the novel GLI module. \textit{When GLI is not used} (only global modeling), the model tends to correlate local regional features, which is consistent with the concept of local population in biological evolution due to geographical constraints, \ie, just like the concept of local search in EA. \textit{With GLI}, explicit local modeling unlocks global modeling potential, forcing global branches to associate more distant features for better results, just as the global/local concept in EA~\cite{mea1,mea2,mea3} that improves performance.}

\subsubsection{$\Delta l$ and $\Delta m$ Distribution of MD-MSA} \label{explanation:mdmsa}
\begin{figure}[!t]
    \centering
    \includegraphics[width=1.0\linewidth]{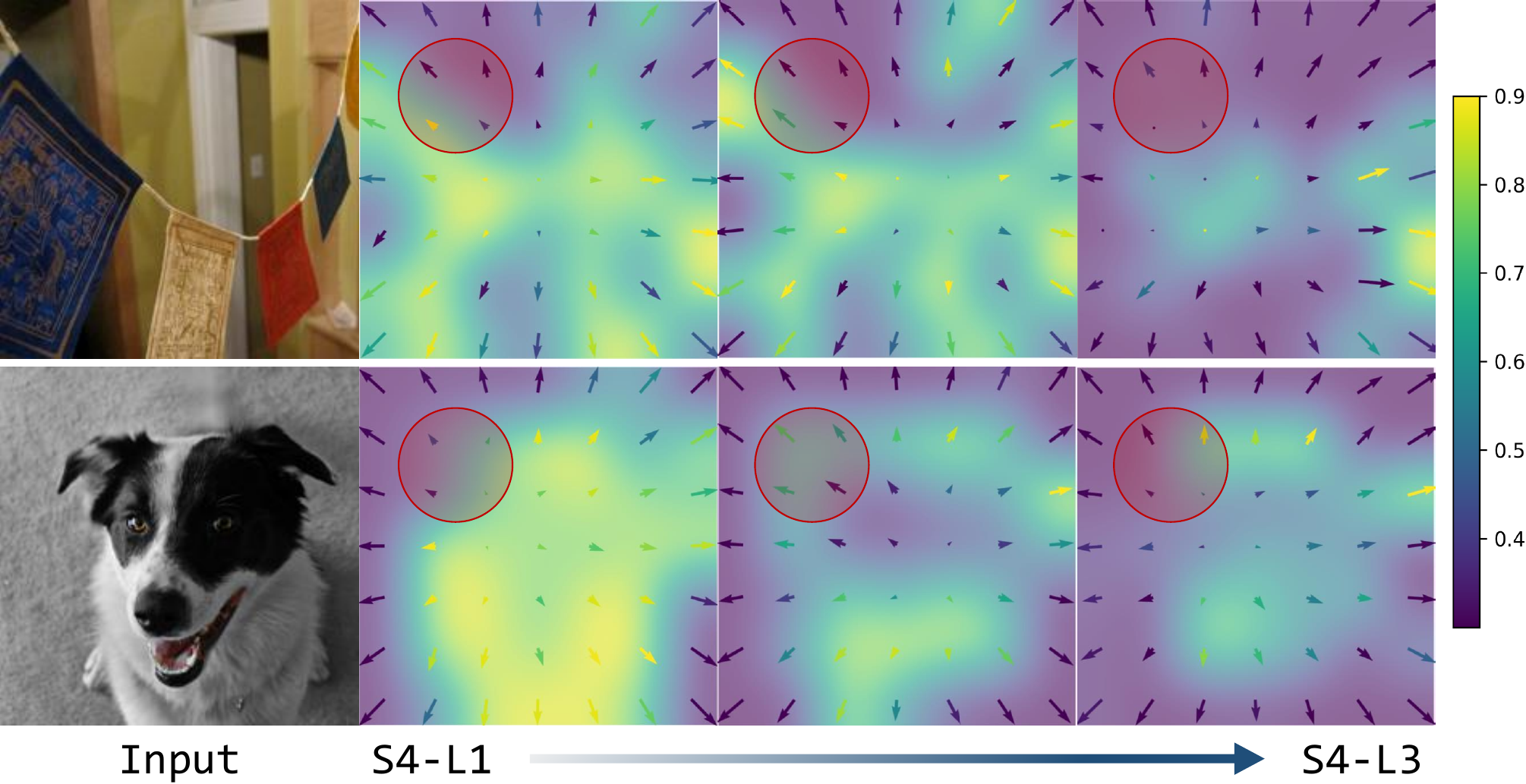}
    \caption{Visualization of deformable offsets (denoted as arrows) and modulation scalar (denoted as color) for MD-MSA of our small model in the last stage. Please zoom in on the red circle for more details.}
    \label{fig:explanation_mdmsa}
\end{figure}

Figure~\ref{fig:explanation_mdmsa} visualizes the learned offset (the longer the arrow, the farther the deformable distance, and the arrow direction indicates sampling direction) and modulation (the brighter the color, the greater the weight) of MD-MSA in stage4. There are differences in offset and modulation of each location in different depths, and the model unexpectedly tends to give more weight to the main object that could describe the main parts of the object. Since we set align\_corners to true when resampling, it has a gradually increasing bias from 0 to 0.5 from the center to the edge. Therefore, the visualization results behave as a whole spreading outwards that may visually weaken changes in each learned position. Please zoom in for better visualization.

\blue{\textbf{Relationship with EA.} Inspired by the irregular spatial distribution among real individuals that are not as horizontal and vertical as the image, we improve the novel MD-MSA module that considers the offset of each spatial position. As show in Figure~\ref{fig:explanation_mdmsa}, different positions (individuals) prefer different offsets and modulation (\ie, direction and scale), just as individuals have different preferences in different regions of the biological world. This modeling method has also been verified in EA, \eg, the improved works~\cite{dae_supp3,dae_supp1,dae_supp2} adopt the similar parameter adaption and feature scaling idea to conduct global feature interaction.}

\subsubsection{Visualization of Attention Map in TRH} \label{explanation:trh}
\begin{figure}[!t]
    \centering
    \includegraphics[width=1.0\linewidth]{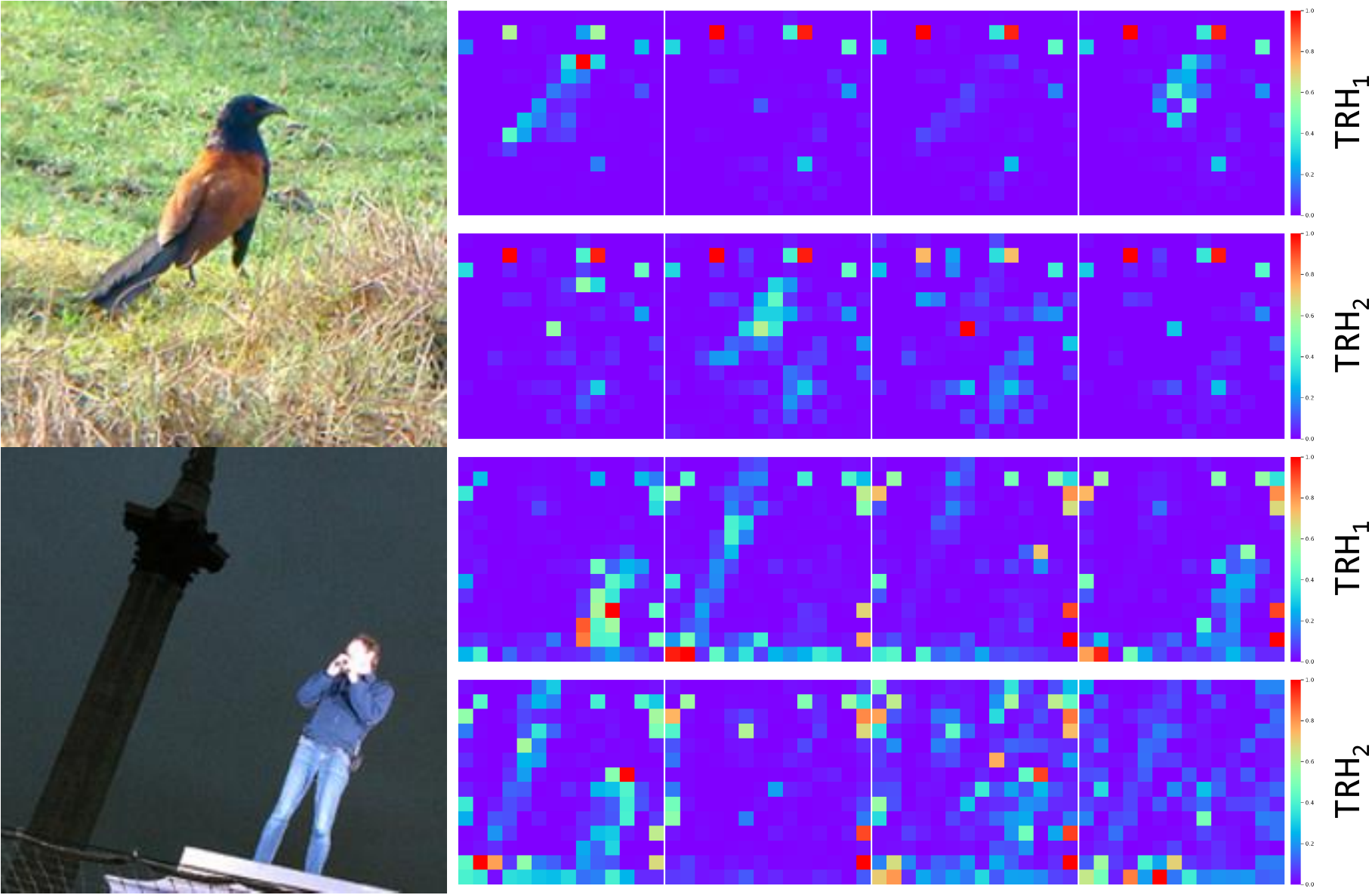}
    \caption{Attention map visualization in TRH for the classification task. The model contains two TRHs, and four head attentions are displayed in each TRH.}
    \label{fig:explanation_trh}
\end{figure}

Taking the classification task as an example, we visualize the attention map in the two-layer TRH that contains multiple heads in the inner cross-attention layer. As shown in Figure~\ref{fig:explanation_trh}, we normalize values of attention maps to [0, 1] and draw them on the right side of the image. Results indicate that different heads focus on different regions, and the deeper TRH$_2$ focuses on a broader area than TRH$_1$ to form the final feature.

\subsubsection{Parameters and FLOPs Distribution} \label{explanation:tiny}
\begin{figure}[!t]
    \centering
    \includegraphics[width=1.0\linewidth]{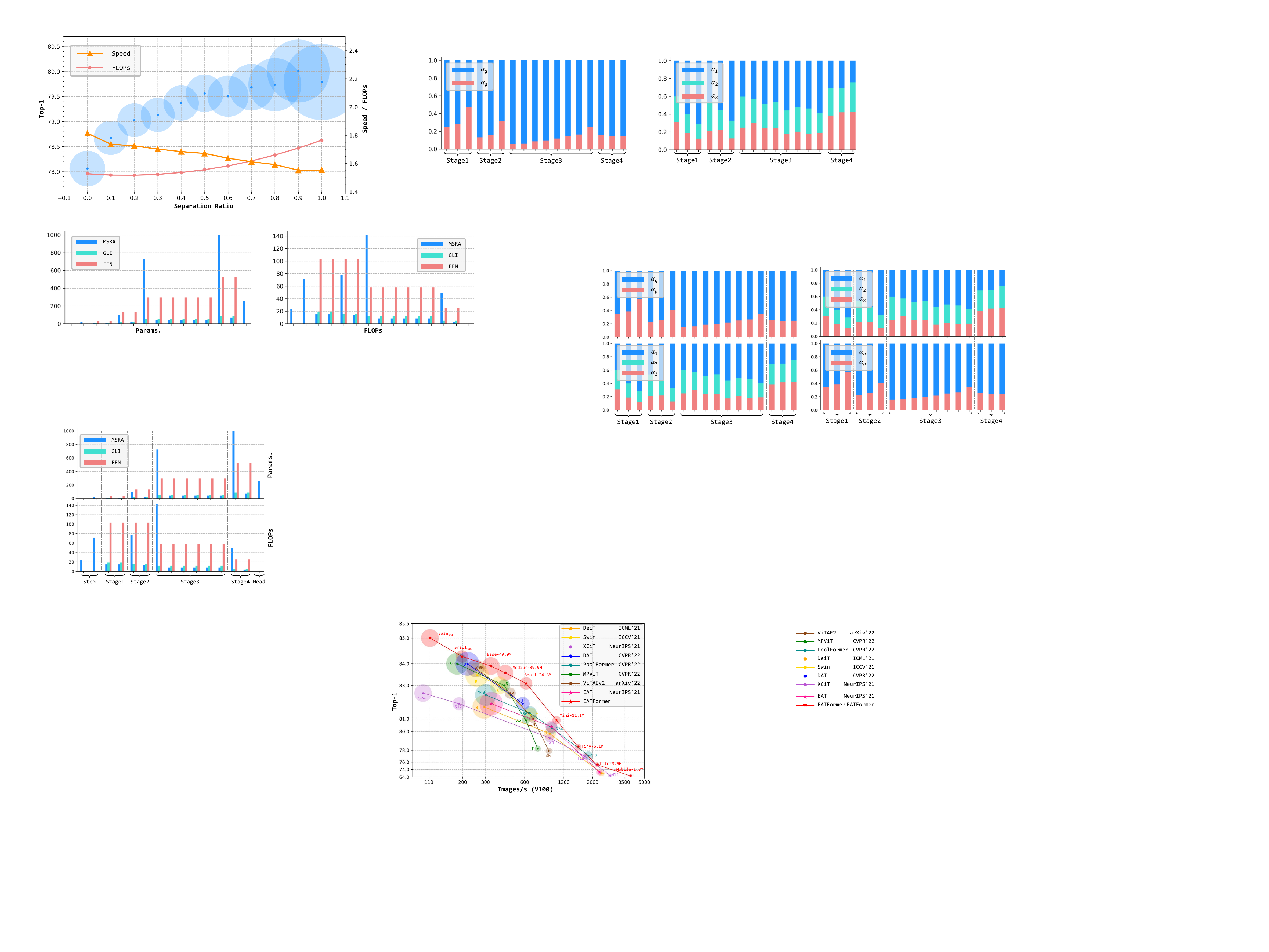}
    \caption{Analyses of Params. and FLOPs distribution.}
    \label{fig:explanation_tiny}
\end{figure}

Taking the designed EATFormer-Tiny as an example, we analyze the distribution of parameters and FLOPs in different layers, where the model contains a stem for resolution reduction, four stages for feature extraction, and a head for target output. As shown in Figure~\ref{fig:explanation_tiny}, the number of parameters is mainly distributed in the deep stage3/4, while FLOPs concentrate in the early stages, and FFN occupies the majority of parameter number calculation. Therefore, we can focus on the optimization of the FFN structure to better balance the comprehensive model efficiency in future work.

\subsubsection{\blue{Works comparison with local/global concepts.}} \label{explanation:localglobal}
\blue{In this paper, the locality in ViT refers to the introduction of CNN with inductive bias into the Transformer structure, and we design the GLI block as a parallel structure that introduces a different local branch beside the global branch. This idea is motivated by some EA variants~\cite{mea1,mea2,mea3} that employ local search procedures besides conventional global search for converging higher-quality solutions. Also, global/local concept is only an idea in the macro sense, and the specific way varies from method to method. \Eg, global/local concept in MPViT~\cite{attn_mpvit} is expressed as parallelism between blocks rather than within each block, while CMT~\cite{cmt} cascades local information into the FFN module rather than the MSA as previous works~\cite{attn_botnet,attn_ceit,efficientformerv2}. Comparatively, our GLI block consists of local convolution and global MD-MSA operations, and Weighted Operation Mixing (WOM) mechanism is further proposed to mix all operations adaptively. So, \textit{we argue that GLI obviously differs comparison methods}. Besides, we compare our method with some contemporary/recent works~\cite{attn_mpvit,cmt,attn_uniformer,attn_vitae2,mobilevit,edgenext,efficientformerv2} which incorporate the global/local concept to their model designs. To further illustrate the differences with these methods, we make comprehensive comparison with them in terms of local/global concepts by several criteria in Table~\ref{table:model_criterion}). Results illustrates the uniqueness of GLI in the technical level.}

\begin{table}[tp]
    \centering
    \caption{\blue{\textbf{Global/local idea comparisons among recent methods}. \textbf{\ding{192}}: Whether the local/global concept is instantiated as a parallel block; \textbf{\ding{193}}: Whether the local operation is combined with MSA; \textbf{\ding{194}}: Whether only one kind of block is used for the model; \textbf{\ding{195}}: Whether feature split is employed for local/global paths; \textbf{\ding{196}}: Whether feature importance is considered for local/global paths. \cmark: Satisfied; \xmark: Unsatisfied.}}
    \label{table:model_criterion}
    \renewcommand{\arraystretch}{1.2}
    \setlength\tabcolsep{6.0pt}
    \resizebox{1.0\linewidth}{!}{
        \begin{tabular}{p{4.6cm}<{\raggedright} p{0.7cm}<{\centering} p{0.7cm}<{\centering} p{0.7cm}<{\centering} p{0.7cm}<{\centering} p{0.7cm}<{\centering}}
            \toprule[0.17em]
            \blue{Method \vs Criterion}                        & \blue{\ding{192}}    & \blue{\ding{193}}    & \blue{\ding{194}}    & \blue{\ding{195}}    & \blue{\ding{196}}    \\
            \hline
            \blue{MPViT}~\cite{attn_mpvit}                     & \blue{\xmark}        & \blue{\xmark}        & \blue{\cmark}        & \blue{\xmark}        & \blue{\xmark}        \\
            \blue{CMT}~\cite{cmt}                              & \blue{\xmark}        & \blue{\xmark}        & \blue{\cmark}        & \blue{\xmark}        & \blue{\xmark}        \\
            \blue{UniFormer}~\cite{attn_uniformer}             & \blue{\xmark}        & \blue{\cmark}        & \blue{\cmark}        & \blue{\xmark}        & \blue{\xmark}        \\
            \blue{ViTAEv2}~\cite{attn_vitae2}                  & \blue{\cmark}        & \blue{\cmark}        & \blue{\xmark}        & \blue{\xmark}        & \blue{\xmark}        \\
            \blue{MobileViT}~\cite{mobilevit}                  & \blue{\xmark}        & \blue{\cmark}        & \blue{\xmark}        & \blue{\xmark}        & \blue{\xmark}        \\
            \blue{EdgeNeXt}~\cite{edgenext}                    & \blue{\xmark}        & \blue{\cmark}        & \blue{\xmark}        & \blue{\xmark}        & \blue{\xmark}        \\
            \blue{EfficientFormerv2}~\cite{efficientformerv2}  & \blue{\xmark}        & \blue{\xmark}        & \blue{\xmark}        & \blue{\xmark}        & \blue{\xmark}        \\
            \hline
            \blue{EATFormer (Ours)}                            & \blue{\cmark}        & \blue{\cmark}        & \blue{\cmark}        & \blue{\cmark}        & \blue{\cmark}         \\
            \toprule[0.12em]
        \end{tabular}
    }
\end{table}

\section{Conclusion} \label{section:con}
This paper explains the rationality of vision transformer by analogy with EA and improves our previous columnar EAT to a novel pyramid EATFormer architecture inspired by effective EA variants. Specifically, the designed backbone consists only of the proposed EAT block that contains three residual parts, \ie, MSRA, GLI, and FFN modules, to model multi-scale, interactive, and individual information separately. Moreover, we propose a TRH module and \emph{improve} an MD-MSA module to boost the effectiveness and usability of our EATFormer further. Abundant experiments on classification and downstream tasks demonstrate the superiority of our approach over SOTA methods in terms of accuracy and efficiency, while ablation and explanatory experiments further illustrate the effectiveness of EATFormer and each analogically designed component.

Nevertheless, we do not use larger models (\eg, >100M), larger datasets(\ie, ImageNet-21K~\cite{imagenet}) or stronger training strategy (\ie, token labeling~\cite{tlt}) for experiments due to limited amount of computation. Also, the architecture recipes are mainly given by our intuition, and the super-parameter could be used to optimize the model structure further. We will explore the above aspects and the combination with self-supervised learning techniques in future works.

\noindent\textbf{Data Availability Statement.} All the datasets used in this paper are available online. ImageNet-1K~\footnote{\url{http://image-net.org}}, COCO 2017~\footnote{\url{https://cocodataset.org}}, and ADE20K~\footnote{\url{http://sceneparsing.csail.mit.edu}} can be downloaded from their official website accordingly. 

\noindent\textbf{Acknowledgement.} This work was supported by a Grant from The National Natural Science Foundation of China(No. 62103363)

\bibliographystyle{spmpsci}
\bibliography{main}
\end{sloppypar}
\end{document}